\documentclass[10pt,twocolumn,letterpaper]{article}

\usepackage{subfiles}
\usepackage{titling}  % if you need separate title pages for the subfiles
\usepackage[pagenumbers]{cvpr} % To force page numbers, e.g. for an arXiv version

\usepackage{fancyhdr}
\usepackage[accsupp]{axessibility}  % Improves PDF readability for those with disabilities.
\usepackage{graphicx}
\usepackage{amsmath}
\usepackage{amssymb}
\usepackage{booktabs}
\usepackage{enumerate}
\usepackage{enumitem}
\usepackage{multirow}
\usepackage[normalem]{ulem}
\useunder{\uline}{\ul}{}

\usepackage[pagebackref,breaklinks,colorlinks]{hyperref}
\newcommand{\midsepremove}{\aboverulesep = 0mm \belowrulesep = 0mm}
\newcommand{\midsepdefault}{\aboverulesep = 0.605mm \belowrulesep = 0.984mm}

\newcommand{\norm}[1]{\left\|#1\right\|} 

\newcommand{\dt}[1]{\textcolor{black}{#1}}

\usepackage[capitalize]{cleveref}
\crefname{section}{Sec.}{Secs.}
\Crefname{section}{Section}{Sections}
\Crefname{table}{Table}{Tables}
\crefname{table}{Tab.}{Tabs.}

\begin{document}

\title{IterativePFN: True Iterative Point Cloud Filtering}

\author{Dasith de Silva Edirimuni$^1$,~Xuequan Lu$^1$\thanks{Corresponding author: X. Lu, supported by PJ03906.PG00507.F002.%.251301
}~,~Zhiwen Shao$^2$\thanks{Z. Shao is supported by the NSFC (No. 62106268).}~,~Gang Li$^1$,~Antonio Robles-Kelly$^{1,4}$,~Ying He$^3$ \\
$^1$School of Information Technology, Deakin University\\
$^2$School of Computer Science and Technology, China University of Mining and Technology \\
$^3$School of Computer Science and Engineering, Nanyang Technological University \\
$^4$Defense Science and Technology Group, Australia \\
\tt\small{\{dtdesilva, xuequan.lu, gang.li, antonio.robles-kelly\}@deakin.edu.au,}
\\
\tt\small{zhiwen\_shao@cumt.edu.cn, yhe@ntu.edu.sg}
}

\maketitle

%%%%%%%%% ABSTRACT
\begin{abstract}
   %Point cloud processing applications range from robotics to urban planning. However, 
   The quality of point clouds is often limited by noise introduced during their capture process. Consequently, a fundamental 3D vision task is the removal of noise, known as point cloud filtering or denoising. State-of-the-art learning based methods focus on training neural networks to infer filtered displacements and directly shift noisy points onto the underlying clean surfaces. In high noise conditions, they iterate the filtering process. However, this iterative filtering is only done at test time and is less effective at ensuring points converge quickly onto the clean surfaces. We propose \textbf{IterativePFN} (iterative point cloud filtering network), which consists of multiple \textbf{IterationModules} that model the true iterative filtering process internally, within a single network. We train our IterativePFN network using a novel loss function that utilizes an adaptive ground truth target at each iteration to capture the relationship between intermediate filtering results during training. This ensures that the filtered results converge faster to the clean surfaces. Our method is able to obtain better performance compared to state-of-the-art methods. The source code can be found at: \url{https://github.com/ddsediri/IterativePFN}. 
\end{abstract}

%%%%%%%%% BODY TEXT
\section{Introduction}
\label{sec:intro}
Point clouds are a natural representation of 3D geometric information and have a multitude of applications in the field of 3D Computer Vision. These applications range from robotics and autonomous driving to urban planning~\cite{Luo-Pillar-Motion,Bekiroglu-PCD-Robotics,Kim-3D-Printing,Urech-Urban-Planning}. They are captured using 3D sensors and comprise of unordered points lacking connectivity information. Furthermore, the capturing of point cloud data is error-prone as sensor quality and environmental factors may introduce noisy artifacts. The process of removing noise is a fundamental research problem which motivates the field of point cloud filtering, also known as denoising. Filtering facilitates other tasks such as normal estimation and, by extension, 3D rendering and surface reconstruction. 

\begin{figure}[!tp]
\centering
\includegraphics[width=0.48\textwidth]{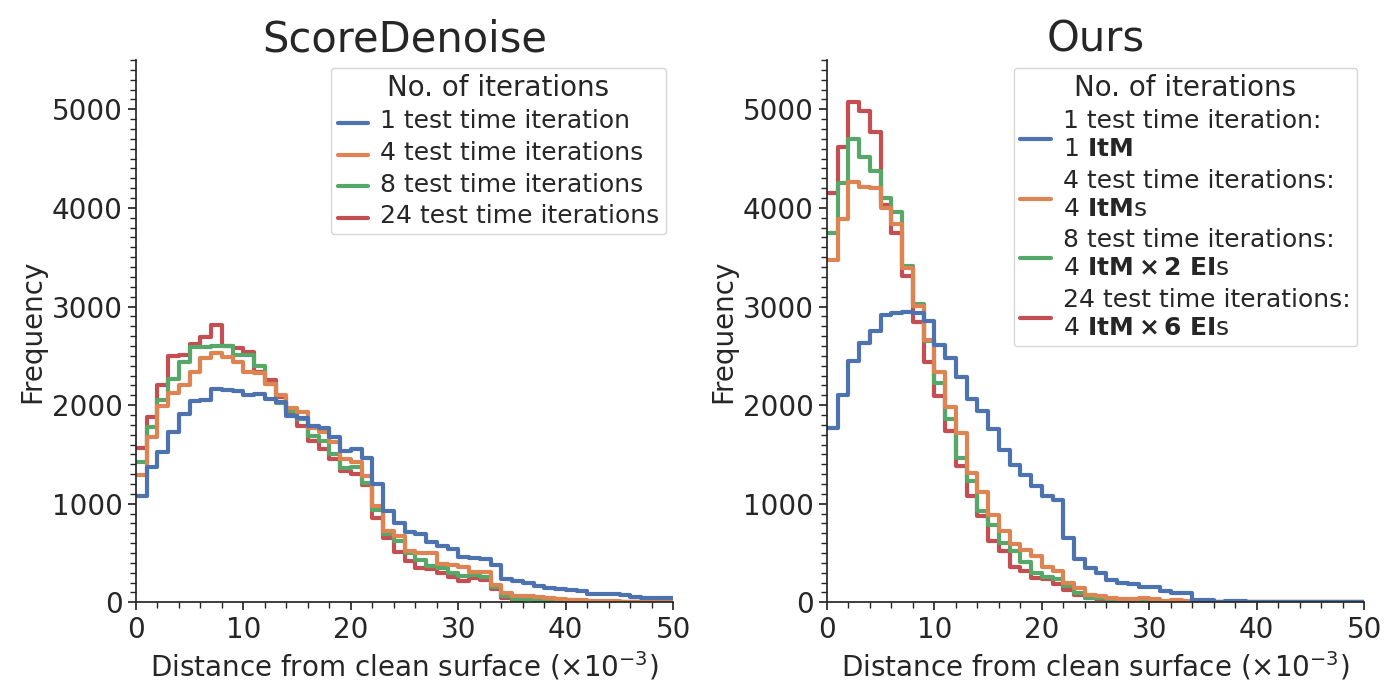}
\caption{Histograms of filtered point distances from clean surface after $1$, $4$, $8$ and $24$ test time iterations for ScoreDenoise~\cite{Luo-Score-Based-Denoising} on the Casting shape with 50K points and 2.5\% Gaussian noise. We compare it with our proposed  IterativePFN where 1 IterationModule (ItM) corresponds to 1 internal iteration and 4 ItMs equal 1 external iteration (EI). There are 4 ItMs in the proposed network. Note 1 ItM is analogous to 1 test time iteration of ScoreDenoise. Our filtering results converge closer to the surface.}
\label{fig:point-dist-high-T}
\end{figure}

\textbf{Conventional} point cloud filtering methods such as MLS based methods~\cite{Alexa--MLS-PSS,APSS-Guennebaud}, bilateral filtering mechanisms~\cite{Digne-Bilateral} and edge recovery algorithms~\cite{Sun-L0,Lu-Low-Rank} rely on local information of point sets, i.e., point normals, to filter point clouds. However, such methods are limited by the accuracy of normals. Alternatives include the Locally Optimal Projection (LOP) family of methods~\cite{Lipman-LOP, Preiner-CLOP, Huang-WLOP}, which downsample and regularize point clouds but incur the loss of important geometric details. More recently, deep learning based filtering methods have been proposed to alleviate the disadvantages and limitations of conventional methods~\cite{Rakotosaona-PCN,Pistilli-GPDNet,Zhang-Pointfilter,Luo-Score-Based-Denoising,Mao-PDFlow}.

\dt{Early \textbf{deep learning} based filtering methods, such as PointProNets~\cite{Roveri-PointProNets}, require pre-processed 2D height maps to filter point clouds. However, the advent of PointNet, PointNet++ and DGCNN, made direct point set convolution a possibility~\cite{Qi-PointNet, Qi-PointNet++, Wang-DGCNN}. Feature encoders based on these architectures were exploited by recent methods to produce richer latent representations of point set inputs and filter noise more effectively~\cite{Rakotosaona-PCN,Luo-DMRDenoise,Zhang-Pointfilter,Luo-Score-Based-Denoising}.} These methods can be broadly characterized as 1) \textbf{resampling}, 2) \textbf{probability} and 3) \textbf{displacement} based methods. Resampling based methods such as DMRDenoise~\cite{Luo-DMRDenoise} suffer from the loss of geometric details as the method relies on identifying downsampled underlying clean surfaces and upsampling along them. ScoreDenoise, which models the gradient-log of the noise-convolved probability to find a point at a given position, iteratively performs Langevin sampling-inspired gradient ascent~\cite{Luo-Score-Based-Denoising} to filter points. However, filtered points are slow to converge to the clean surface after many test time iterations of filtering, as illustrated in Fig.~\ref{fig:point-dist-high-T}. By contrast, for an IterativePFN network with 4 IterationModules, where 1 iterationModule (ItM) represents 1 internal iteration of filtering and is analogous to 1 test time iteration of ScoreDenoise, we see that a higher number of filtered points converge closer to the clean surface within the same number of test time iterations.

\begin{figure*}[!tp]
\centering
\includegraphics[width=0.98\linewidth]{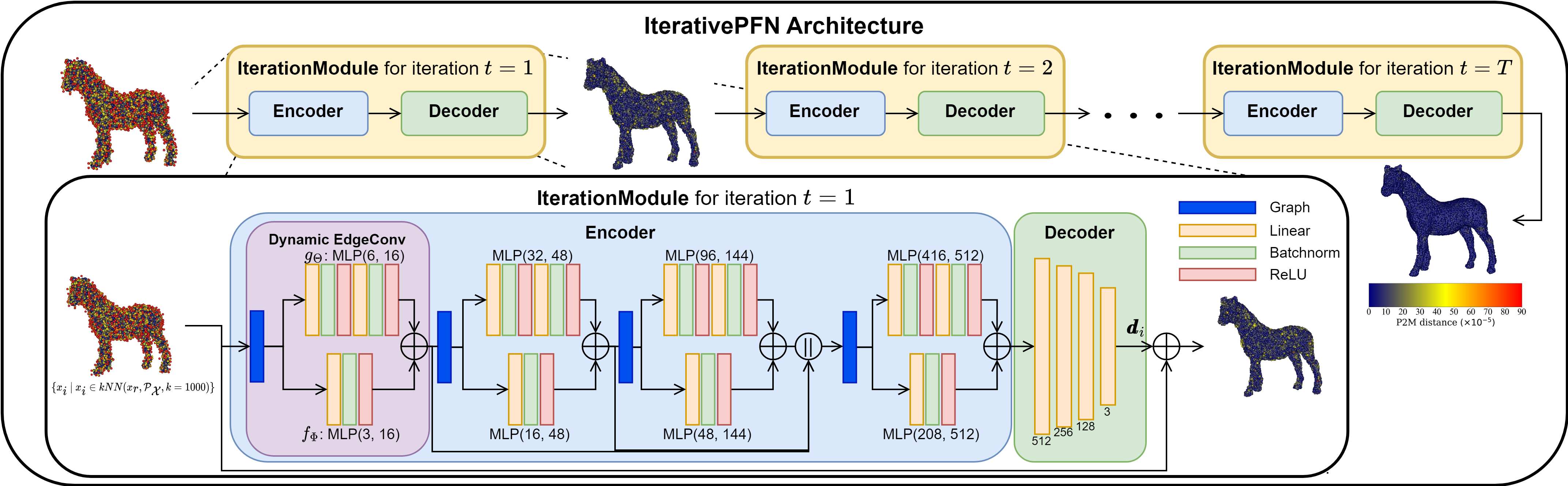}
\caption{Overview of our IterativePFN method. Unlike existing methods which need to be iteratively applied during test-times, we explicitly model $T$ iterations of filtering in our network using $T$ IterationModules.}
\label{fig:iterativepfn-network}
\end{figure*}

Among displacement based methods, PointCleanNet (PCN)~\cite{Rakotosaona-PCN} shows  sensitivity to high noise while Pointfilter~\cite{Zhang-Pointfilter} utilizes a bilateral filtering inspired weighted loss function that causes closely separated surfaces to collapse into a single surface during filtering. Moreover, gradient ascent and displacement based methods filter point clouds iteratively during test times and do not consider true iterative filtering during training. %In general, they do not guarantee convergence to the clean surfaces. 
Although RePCDNet~\cite{Chen-RePCD} offers a recurrent neural network inspired alternative to capture this information during training, at each iteration RePCDNet attempts to directly shift points onto the clean surface without considering, that at different iterations, their existing residual noise, in decreasing order w.r.t. iteration number. Furthermore, it uses a single network to filter points, increasing the burden on the network to correctly distinguish between noise scales and requires multiple test time iterations which lead to low efficiency. 
Based on these major limitations, we propose:
\begin{itemize}
    \item a novel neural network architecture of stacked encoder-decoder modules, dubbed \textbf{IterationModule}, to model the true iterative filtering process internally (see Fig.~\ref{fig:iterativepfn-network}). Each IterationModule represents an iteration of filtering and the output of the $\tau$-th IterationModule becomes the input for the $\tau+1$-th IterationModule. Thereby, the $\tau+1$-th IterationModule represents the filtering iteration $t=\tau+1$. This allows the network to develop an understanding of the filtering relationship across iterations.
    \item a novel loss function that formulates the nearest neighbor loss at each iteration as the $L_2$ norm minimization between the filtered displacements, inferred by the $\tau$-th IterationModule, and the nearest point within a target point cloud at $t=\tau$, of a lower noise scale $\sigma_\tau$ compared to the noise scale $\sigma_{\tau-1}$ of the target at $t=\tau-1$. This promotes a gradual filtering process that encourages convergence to the clean surface.
    \item a generalized patch-stitching method that designs Gaussian weights when determining best filtered points within overlapping patches. Patch stitching improves efficiency as it facilitates filtering multiple points simultaneously. %A minor contribution is the generalization of patch-stitching, proposed in~\cite{Zhou-Patch-Stitching}, to function independently of network inferred weights.
\end{itemize}

We conduct comprehensive experiments, in comparison with state-of-the-art methods, which demonstrate our method's advantages on both synthetic and real world data.

\section{Related Work}
\label{sec:rel-work}
\textbf{Traditional methods}. Moving Least Squares (MLS) introduced by Levin~\cite{MLS-Levin} was extended to tackle point cloud filtering by the work of Alexa et al.~\cite{Alexa--MLS-PSS} with the intent of minimizing the approximation error between filtered surfaces and their ground truth. Thereafter, Adamson and Alexa proposed Implicit Moving Least Squares (IMLS)~\cite{IMLS-Adamson} and Guennebaud and Gross introduced Anisotropic Point Set Surfaces (APSS)~\cite{APSS-Guennebaud}, where both methods attempted to enhance the filtering performance on point clouds with sharp geometric features. However, these MLS techniques rely on parameters of the local surface, such as point normals. In order to alleviate this, Lipman et al.~\cite{Lipman-LOP} proposed the Locally Optimal Projection (LOP) method. In addition to being independent of local surface parameters, LOP based methods downsample and regularize the input point clouds. This work spawned a new class of point cloud filtering methods, including Continuous Locally Optimal Projection (CLOP) by Preiner et al.~\cite{Preiner-CLOP} and Weighted Locally Optimal Projection (WLOP) by Huang et al.~\cite{Huang-WLOP}. In general, both classes of methods suffer from the same disadvantage: they are unable to preserve sharp geometric features of the underlying clean surfaces and are sensitive to noise.

Moreover, Cazals and Pouget proposed fitting n-order polynomial surfaces to point sets~\cite{Cazals-Jet} to calculate quantities of interest such as point normals and curvature and has seen learning based applications in~\cite{Ben-Shabat-DeepFit,Zhu-AdaFit}. Moreover, by projecting noisy points onto the fitted surface, their filtered counterparts can be obtained. Digne introduced a similarity based filtering method that uses local descriptors for each point and, subsequently, exploits the similarity between descriptors to determine filtered displacements at each point~\cite{Digne-Similarity}. More recently, a generalization of the mesh bilateral filtering mechanism~\cite{Fleishman-Bilateral} for point clouds, was proposed by Digne and de Franchis that filters points based on an anisotropic weighting of point neighborhoods. This weighting considers both point positions and their normals. Other methods such as the $L_1$ and $L_0$ minimization methods of Avron et al. and Sun, Schaefer, and Wang~\cite{Avron-L1, Sun-L0} and the Low Rank Matrix Approximation of Lu et al.~\cite{Lu-Low-Rank} estimate normals that are used in filtering algorithms to denoise point clouds. The filtering process has also been reformulated as a sparse optimization problem which can be solved by Augmented Lagrangian Multipliers, in the work of Remil et al.~\cite{Remil-Data-Driven-Sparse-Priors}.

\textbf{Deep learning based methods}.
Among learning based methods, PointProNets was proposed by Roveri et al.~\cite{Roveri-PointProNets} and utilized a traditional CNN architecture to filter noisy 2D height maps which were re-projected into 3D space. By contrast, the Deep Feature Preserving (DFP) network of Lu et. al used a traditional CNN with 2D height maps to estimate point normals that can be used in conjunction with the position update algorithm of~\cite{Lu-Low-Rank}. With the advent of PointNet and PointNet++~\cite{Qi-PointNet,Qi-PointNet++}, direct convolution of point sets became possible. Furthermore, Wang et. al~\cite{Wang-DGCNN} proposed a graph convolutional architecture which consumed nearest-neighbor graphs generated from point sets to produce rich feature representations of points. The following methods use PointNet or DGCNN inspired backbones for processing point cloud inputs. Yu et al. developed EC-Net to perform denoising by upsampling in an edge-aware manner and, thereby, consolidate points along a surface~\cite{Yu-EC-Net}. A similar approach is adopted by Luo and Hu in their DMRDenoise mechanism~\cite{Luo-DMRDenoise} which sequentially identifies an underlying downsampled, less noisy, manifold and subsequently upsamples points along it. PCN, by Rakotosaona et al.~\cite{Rakotosaona-PCN}, filters central points within point cloud patches. PCN infers these filtered displacements by utilizing a loss which minimizes the $L_2$ norm between predicted displacements and points within the ground truth patch. Furthermore, they add a repulsion term  to ensure a regular distribution of points. Zhang et al. proposed Pointfilter~\cite{Zhang-Pointfilter} that develops this displacement based line of inquiry. They use ground truth normals during training to calculate a bilateral loss. Pistilli et al. proposed GPDNet, a graph convolutional architecture to infer filtered displacements~\cite{Pistilli-GPDNet}. Recently, Chen et al. proposed RePCD-Net, a recurrent architecture that iteratively filters point clouds during training and considers the training losses at each iteration. However, this method must still be applied iteratively during test time~\cite{Chen-RePCD}.

Langevin sampling of noisy data to iteratively recover less noisy data, at test-time, was initially proposed in the 2D generative modelling field~\cite{Song-Score-Matching} and has been applied to point cloud generation and reconstruction~\cite{Cai-ShapeGF,Luo-Diffusive}. This relies on learning the unnormalized gradient-log probability distribution of the data. Luo and Hu extended this to point cloud filtering. Their method, ScoreDenoise, models the gradient-log of the underlying noise convolved probability distribution for point cloud patches~\cite{Luo-Score-Based-Denoising}. Mao et al. proposed PDFlow to uncover latent representations of noise-free data at higher dimensions by utilizing normalizing flows and, thereafter, obtain the filtered displacements~\cite{Mao-PDFlow}.

\section{Research Problem and Motivation}
\label{sec:problem-motivation}
As discussed in Sec.~\ref{sec:intro}, previous point cloud filtering methods can be characterized as resampling, probability and displacement based methods, the second of which includes score-based methods. We are motivated by the interplay between filtering objectives of displacement and score-based methods and how a score-based method can inform the construction of a truly iterative displacement based method, that is iterative in its training objective. \dt{PCN~\cite{Rakotosaona-PCN} proposed the idea that the filtering objective should aim to regress a noisy point $\pmb{x}_i$ back onto the underlying clean surface using a displacement $\pmb{d}_i$. This displacement is the output of the regression network, which takes a noisy point cloud patch $\mathcal{X}$, centered at $\pmb{x}_i$, as its input. The filtering objective is expressed as} $\Tilde{\pmb{x}}_i = \pmb{x}_i + \pmb{d}_i$ where $\Tilde{\pmb{x}}_i$ is the filtered point. During testing, the output at time $t=\tau$ is taken as input for the next iteration at time $t=\tau+1$. This leads to the test-time iterative filtering objective:
\begin{align}
\label{eq:pcn-position-update}
    \Tilde{\pmb{x}}^{(t)}_i = \Tilde{\pmb{x}}^{(t-1)}_i + \pmb{d}^{(t)}_i, t = 1,\cdots,T
\end{align}
PCN's training objective is motivated by the need to regress the noisy point back to the clean patch $\mathcal{Y}$, while ensuring it is centered within it. Thus, the PCN training objective is achieved by the following loss:
\begin{align}
\label{eq:pcn-loss}
L^{PCN}_i =~&\alpha\min_{\pmb{x}_j \in \mathcal{Y}}\norm{\pmb{d}_i - \pmb{\delta x}_j}^2_2 + (1-\alpha)\max_{\pmb{x}_j \in \mathcal{Y}}\norm{\pmb{d}_i - \pmb{\delta x}_j}^2_2,
\end{align}
where $\pmb{\delta x}_j=\pmb{x}_j-\pmb{x}_i$. The score-based method, ScoreDenoise~\cite{Luo-Score-Based-Denoising}, has a similar filtering objective given by:
\begin{align}
\label{eq:luo-position-update}
    \Tilde{\pmb{x}}^{(t)}_i = \Tilde{\pmb{x}}^{(t-1)}_i + \alpha^{(t)}\mathcal{E}_i(\Tilde{\pmb{x}}^{(t-1)}_i), ~t = 1,\cdots,T
\end{align}
where $\mathcal{E}_i(\pmb{x}) = (1/K)\sum_{\pmb{x}_j \in kNN(\pmb{x}_i)}\mathcal{S}_j(\pmb{x})$ is an ensemble average of scores for the $k$-neighbors of the given point. %These scores $\mathcal{S}_i(\pmb{x})=\text{Score}(\pmb{x}-\pmb{x}_i, h_i)$ are predicted by the ScoreDenoise network, where $h_i$ is the feature vector for $\pmb{x}_i$. 
These scores are predicted by the ScoreDenoise network and correspond to the gradient-log of the noise-convolved probability $\nabla_{\pmb{x}}\log[(p*n)(\pmb{x})]$. Its training objective is:
\begin{align}
\label{eq:luo-loss}
    L^{SD}_i = \mathbb{E}_{\pmb{x}\sim\mathcal{N}(\pmb{x}_i)}\left[\norm{s(\pmb{x})-\mathcal{S}_i(\pmb{x})}^2_2\right],
\end{align}
where $s(\pmb{x})=NN(\pmb{x}, \mathcal{Y})-\pmb{x}$, and $NN(\pmb{x}, \mathcal{Y})$ is the nearest point to $\pmb{x}$ in the clean patch.

Based on the above analysis, we note the training objective of~\cite{Luo-Score-Based-Denoising} considers a concentration of neighbors $\pmb{x}$ near $\pmb{x}_i$ while PCN's training objective only attempts to infer the displacement for $\pmb{x}_i$. Despite the differences of these two methods, both groups of methods perform reasonably well at removing noise. This is in part due to the fact that during the test phase they emulate the Langevin equation of a reverse Markov process that iteratively removes noise. However, we also observe several drawbacks: 
\textbf{Firstly}, both displacement and score-based methods are iterative only at test time. During training, they see noisy patches with underlying Gaussian noise distributions and attempt to infer the filtered displacements~\cite{Rakotosaona-PCN} or scores~\cite{Luo-Score-Based-Denoising} to shift points directly onto the clean patch. Consequently, the PCN position update, Eq.~\eqref{eq:pcn-position-update}, and gradient ascent equation, Eq.~\eqref{eq:luo-position-update}, both neglect the fact that filtered points have an additive noise contribution $\sigma_\tau\xi$ where $\xi\sim\mathcal{N}(0,I)\land\sigma_\tau\in\mathbb{R}$ which is due to the stochastic nature of the Langevin equation.

\begin{figure}[!tp]
\centering
\includegraphics[width=\linewidth]{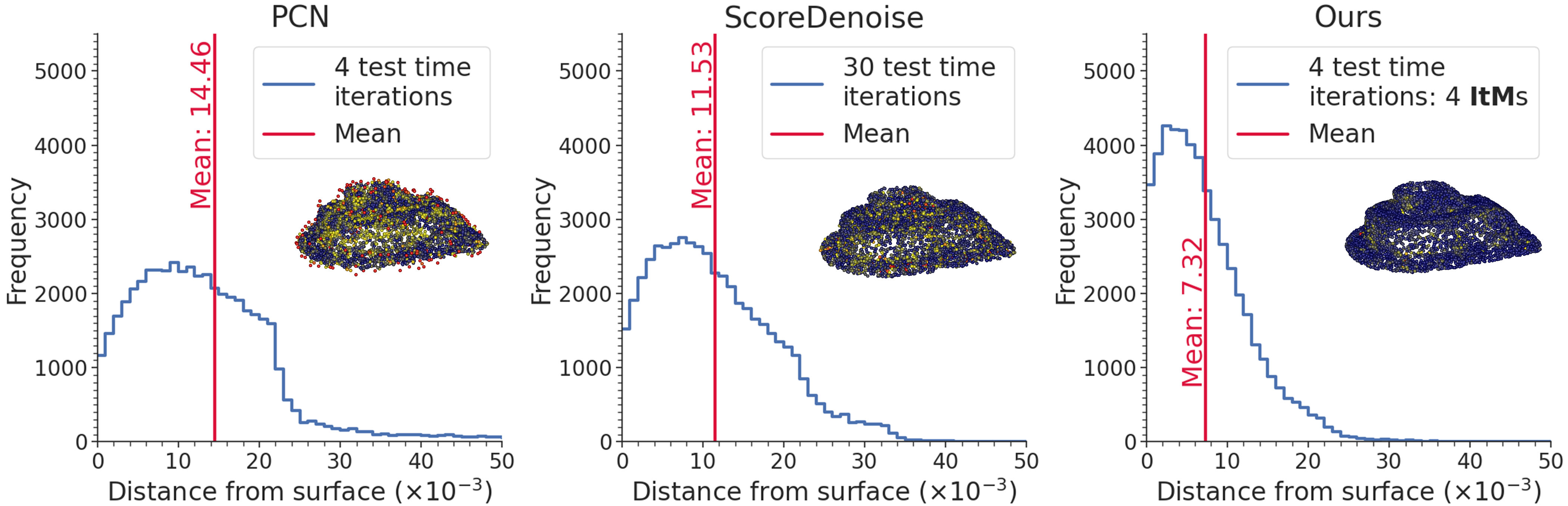}
\caption{Histograms of distance to clean surface for each filtered point cloud. We look at the initial Casting shape at 50K point resolution and Gaussian noise of 2.5\%. }
\label{fig:point-distribution}
\end{figure}

\textbf{Secondly}, ScoreDenoise~\cite{Luo-Score-Based-Denoising} relies on a decaying step size, $\alpha^{(t)}$, to ensure the convergence of Eq.~\eqref{eq:luo-position-update} to a steady state. However, this can indeed be very slow to converge\cite{Cai-ShapeGF}. These observations motivate us to propose a learning based filtering method that effectively incorporates iterative filtering into the training objective such that the network is able to successfully recognize the underlying noise distribution at each iteration and, thereby, impose a more robust filtering objective wherein filtered results converge faster to the clean surface. Fig.~\ref{fig:point-distribution} depicts histograms of the number of filtered points at a given distance from the clean surface. We use the default test time iteration numbers for PCN and ScoreDenoise, which are 4 and 30, respectively. For our method, we use 4 IterationModules, i.e., \textit{4 internal iterations} or \textit{1 external iteration}. For PCN, the mean distance from the clean surface is 0.014 (in units w.r.t. the bounding sphere's radius) while for ScoreDenoise it is 0.012 which are both much higher than the mean value of our method: 0.007. For point clouds filtered using PCN and ScoreDenoise, larger numbers of filtered points lie further away from the underlying clean surfaces after applying multiple iterations of filtering at test time but our method filters points closer to the surface after a single external iteration.

\section{True Iterative Point Cloud Filtering}
\label{sec:tuly-iterative-filtering}
In this section we look in detail at our proposed \textbf{IterativePFN} network that consists of a stack of encoder-decoder pairs. Each pair, dubbed \textbf{IterationModule}, models a single internal filtering iteration and corresponds to a test time iteration of a method such as ScoreDenoise or PCN. At each internal iteration, the corresponding IterationModule takes a patch of noisy points and constructs a directed graph. The patch, and its associated graph, are then consumed by Dynamic EdgeConv layers to generate latent representations of vertices for regressing filtered displacements for each patch point. To model the effect of $T$ iterations, we use $T$ IterationModules. An external iteration represents the effect of filtering a point cloud using all IterationModules in the network. During inference, we use farthest point sampling on the noisy point cloud to generate input patches. However, this implies overlapping regions between different patches. Therefore, we propose a generalized method of patch stitching to recover the best filtered points.

\subsection{Graph convolution and network architecture}
\label{subsec:graph-gen-network-arch}
Graph convolution based neural networks have been shown to generate rich latent representations of 3D point based graphs~\cite{Wang-DGCNN, Pistilli-GPDNet}. In particular, the work of Pistilli et al.~\cite{Pistilli-GPDNet} focuses on graph convolution for filtering point clouds. Apart from generating rich feature representations, graph convolution also provides an advantage on efficiency and lower inference times over point set convolution methods, that use PointNet inspired backbones, such as PCN and Pointfilter. These methods are designed to consume patches of points to filter a single central patch point, which leads to slower processing. In graph convolution based methods, all vertices within the graph are simultaneously processed, with filtered displacements being inferred for each single vertex (point) within the graph. For our IterativePFN network, we use a modified form of the Dynamic EdgeConv layers, proposed by Wang et al.~\cite{Wang-DGCNN}, within each encoder, to generate rich feature representations of the input graphs. \dt{Formally, the vertex feature is updated as $ \mathbf{h}^{l+1}_i = f_{\mathbf{\Phi}}(\mathbf{h}^{l}_i) + \sum_{j:(i,j) \in \mathcal{E}} g_{\mathbf{\Theta}}(\mathbf{h}^{l}_i \, \Vert \, \mathbf{h}^{l}_j - \mathbf{h}^{l}_i)$,} where $i$ is a vertex on the graph, $(i,j)$ form an edge and $(*\,\Vert\,*)$ represents concatenation. Here, $f_{\mathbf{\Phi}}:\mathbb{R}^{F^l}\rightarrow\mathbb{R}^{F^{l+1}}$ and $g_{\mathbf{\Theta}}:\mathbb{R}^{F^l}\times\mathbb{R}^{F^l}\rightarrow\mathbb{R}^{F^{l+1}}$ are parametrized by MLPs. $F^{l}$ is the feature dimension at layer $l$. Thereafter, decoders consisting of 4 Fully Connected (FC) layers, consume these latent features to infer displacements at each iteration.       

\subsection{Graph construction}
\label{subsec:graph-construction}
Given a clean point cloud $\mathcal{P}_\mathcal{Y} = \{ \pmb{y}_i\ |\ \pmb{y}_i \in \mathbb{R}^3,\ i\ = 1, ..., n \}$, perturbed points are formed by adding Gaussian noise with standard deviation $\sigma_0$ between 0.5\% and 2\% of the radius of the bounding sphere. Therefore, a noisy point cloud is given by \dt{$\mathcal{P}_\mathcal{X} = \mathcal{P}_\mathcal{Y} + \sigma_0\xi,~\xi\sim\mathcal{N}(0, I)$.} Subsequently, given a reference point $\pmb{x}_r \in \mathcal{P}_\mathcal{X}$, we obtain patches $\mathcal{X}$ such that $\mathcal{X} = \{\pmb{x}_i\ |\ \pmb{x}_i \in \mathcal{P}_\mathcal{X} \land \pmb{x}_i \in \dt{kNN(\pmb{x}_r, \mathcal{P}_\mathcal{X}, k=1000)} \}$ where \dt{$kNN(\pmb{x}, \mathcal{P}_\mathcal{X}, k=m)$} refers to the $m$ nearest neighbors of the point $\pmb{x}$ in the point set $\mathcal{P}_\mathcal{X}$. Specifically, we select the 1000 nearest neighbors of $\pmb{x}_r$. The corresponding clean patch $\mathcal{Y}$ w.r.t. $\mathcal{X}$ is given by $\mathcal{Y} = \{\pmb{y}_i\ |\ \pmb{y}_i \in \mathcal{P}_\mathcal{Y} \land \pmb{y}_i \in \dt{kNN(\pmb{x}_r, \mathcal{P}_\mathcal{Y}, k=1200)} \}$. Boundary points of $\mathcal{X}$ may originate from points on the clean surface outside the 1000 nearest neighbors of $\pmb{x}_r$ in $\mathcal{Y}$, that are perturbed onto $\mathcal{X}$ due to noise. To account for this, we follow ScoreDenoise's strategy by setting the number of points in $\mathcal{Y}$ to be slightly greater than $\mathcal{X}$. \dt{Thereafter, patches are centered at the reference point, i.e., $\mathcal{X}=\mathcal{X}-\pmb{x}_r$ and $\mathcal{Y}=\mathcal{Y}-\pmb{x}_r$. Patch point indices are then used to construct a directed graph $\mathcal{G}=(\mathcal{V}, \mathcal{E})$ of vertices and edges where $\mathcal{V} = \{i\ |\ \pmb{x}_i \in \mathcal{X} \}$ and $ \mathcal{E} = \{(i, j)\ |\ \pmb{x}_i, \pmb{x}_j \in \mathcal{X} \land \pmb{x}_j \in \dt{kNN(\pmb{x}_i, \mathcal{X}, k=32) \}}$. For each vertex, edges are formed with its 32 nearest neighbors.}

\subsection{Displacement based iterative filtering during training}
\label{subsec:displacement-iterative-mechanism}

\begin{figure}[!tp]
\centering
\includegraphics[width=0.9\linewidth]{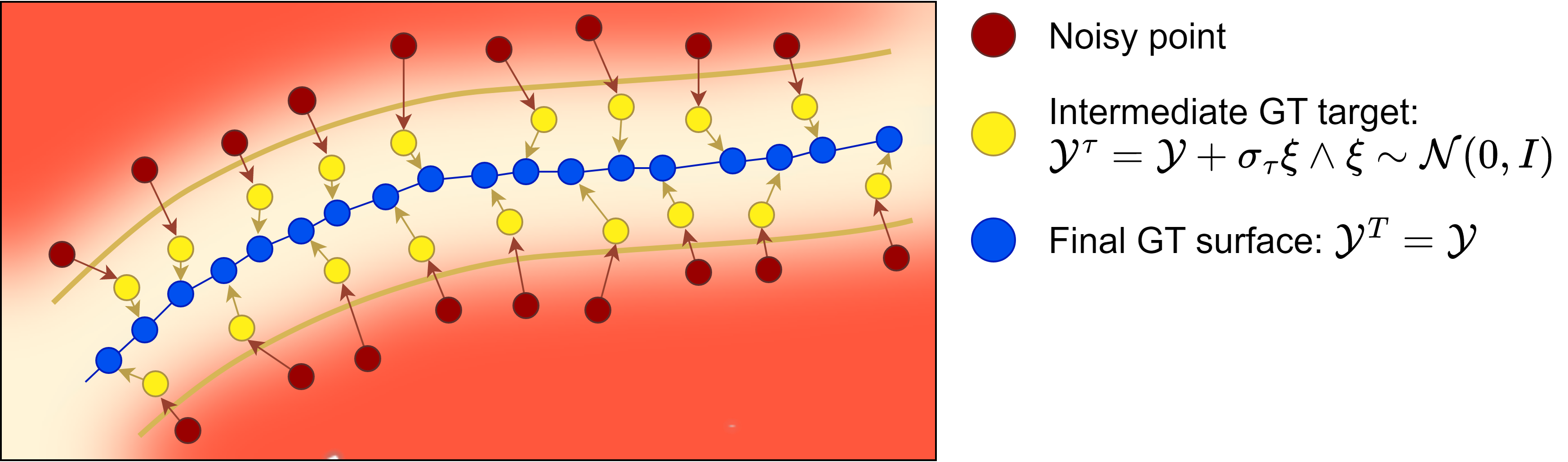}
\caption{An adaptive ground truth target is used during intermediate filtering iterations to encourage convergence to the surface. }
\label{fig:filtering-step-by-step}
\end{figure}

While previous displacement based methods, such as PCN and Pointfilter, attempt to infer the filtered displacement required to directly shift a noisy point onto the clean patch, our method uses multiple IterationModules to iteratively reduce noise. %Inspired by the displacement based training objective of~\cite{Rakotosaona-PCN}, 
Thus we propose a novel training objective aimed at gradually reducing point cloud noise as illustrated in Fig.~\ref{fig:filtering-step-by-step}. Our loss function is given by,
\dt{
\begin{align}
\label{eq:initial-nn-loss}
L^{(\tau)}_i(\mathcal{Y}^{(\tau)}) =~&\norm{\pmb{d}^{(\tau)}_i - \pmb{\delta x}^{(\tau)}_i(\mathcal{Y}^{(\tau)})}^2_2,
\end{align}
where $\pmb{d}^{(\tau)}_i$, the filtered displacement, is the output of IterationModule $t=\tau$, \dt{$\pmb{\delta x}^{(\tau)}_i(\mathcal{Y}^{(\tau)}) = NN(\pmb{x}^{(\tau-1)}_i, \mathcal{Y}^{(\tau)})-\pmb{x}^{(\tau-1)}_i$} and $NN(\pmb{x}^{(\tau-1)}_i, \mathcal{Y}^{(\tau)})$ is the nearest neighbor of $\pmb{x}^{(\tau-1)}_i$ in $\mathcal{Y}^{(\tau)}$. Moreover, $\mathcal{Y}^{(\tau)}=\mathcal{Y}+\sigma_\tau\xi~\land~\xi\sim\mathcal{N}(0, I)$ is the \textbf{adaptive ground truth} patch at iteration $t=\tau$.} This target patch contains less noise compared to $\mathcal{Y}^{(\tau-1)}$. We ensure $\sigma_0>\sigma_1>\cdots>\sigma_{T}$ where we set $\sigma_T=0$. That is, $\mathcal{Y}^{(T)}=\mathcal{Y}$. For simplicity, \dt{we set $\sigma_{\tau+1} = \sigma_{\tau}/\delta$, where $\delta = 16/T$} is the decay hyperparameter controlling the noise scale of the target patch, \dt{$12\geq T\geq 2~\text{and}~T-2\geq\tau\geq 0$.} More precisely, we distinguish our training objective from PCN and Pointfilter by minimizing the $L_2$ loss between the filtered displacements and the nearest neighbor in a less noisy version of the same, underlying, clean surface.

\subsection{Generalized patch stitching}
\label{subsec:gen-patch-stitching}

\begin{figure}[!ht]
\centering
\includegraphics[width=0.8\linewidth]{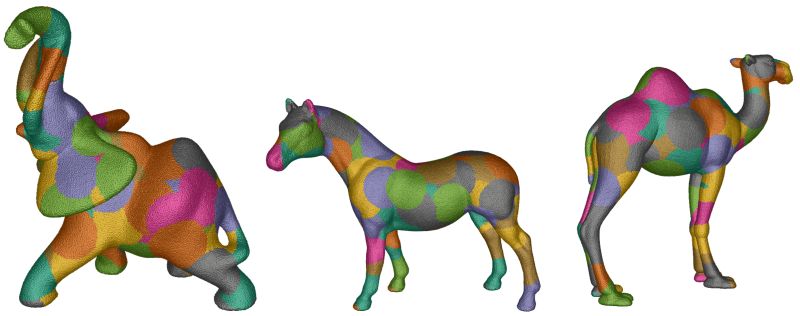}
\caption{Patches of nearest neighbor points constructed using farthest point sampling along the surface.}
\label{fig:multi-patch}
\end{figure}

During inference, we use farthest point sampling to obtain $R$ reference points $\{\pmb{x}_r\}^{R}_{r=1}$ and construct input patches using these points. All points within a patch are filtered simultaneously, as opposed to methods such as PCN and Pointfilter where a given patch is used to filter only a single central point. As shown in Fig.~\ref{fig:multi-patch}, patches may have overlapping regions with repeated points. Therefore, we must find the patches that optimally filter these repeated points. Zhou et al., proposed patch stitching as a mechanism to obtain the best filtered points within such regions\cite{Zhou-Patch-Stitching}. However, this method relies on network inferred weights, obtained using a self-attention module, to infer the best patch that yields the best filtered point. However, this is computationally expensive as it relies on a self-attention module and is not easy to generalize to methods that do not use such a module. Hence, we are motivated to design a more general patch stitching method that can be used in conjunction with any graph based neural network architecture. We observe that filtering results of points near the boundary of patches are less favorable than those closer to the reference points, $\pmb{x}_r$, which are located at the center. This is due to the fact that points close to the patch boundary have asymmetric neighborhoods and, as such, receive biased signals during graph convolution that affect their filtered output. \dt{Therefore, an intuitive strategy is to weight input points based on their proximity to the reference point $\pmb{x}_r$ according to a Gaussian distribution,
$w_i = \frac{\exp\left(-\norm{\pmb{x}_i-\pmb{x}_r}^2_2/r^2_s\right)}{\sum_i \exp\left(-\norm{\pmb{x}_i-\pmb{x}_r}^2_2/r^2_s\right)}$, where $r_s$ is the support radius, which we empirically set to $r_s=r/3$ with $r$ being the patch radius. }

\subsection{\dt{Iterative filtering and patch stitching}}

\dt{Now, the loss at $t=\tau$ is given by $L^{(\tau)} = \sum_i w_i L^{(\tau)}_i$,
% \begin{align}
%     L^{(\tau)} = \sum_i w_i L^{(\tau)}_i
% \end{align}
where the individual loss contribution at each point is given by $L^{(\tau)}_i$, according to Eq.~\eqref{eq:initial-nn-loss}. Finally, we sum loss contributions across iterations to obtain the final loss,}
\begin{align}
    \label{eq:initial-nn-loss-full}
    \mathcal{L}_a = \sum^{T}_{\tau=1} L^{(\tau)},
\end{align}
which is used to train our network. It enjoys the distinction of being a truly iterative train time filtering solution that, at test times, can consume noisy patches and filter points without needing to resort to multiple external iterations. 

\midsepremove
\begin{table*}[!tp]
\setlength{\tabcolsep}{5pt}
\centering
\begin{tabular}{l|llllll|llllll}
\toprule
\multicolumn{1}{c|}{\multirow{3}{*}{Method}} & \multicolumn{6}{c|}{10K points} & \multicolumn{6}{c}{50K points} \\
\cmidrule{2-13}
 & \multicolumn{2}{c|}{1\% noise} & \multicolumn{2}{c|}{2\% noise} & \multicolumn{2}{c|}{2.5\% noise} & \multicolumn{2}{c|}{1\% noise} & \multicolumn{2}{c|}{2\% noise} & \multicolumn{2}{c}{2.5\% noise} \\
\cmidrule{2-13}
 & CD & P2M & CD & P2M & CD & P2M & CD & P2M & CD & P2M & CD & P2M \\
\midrule
Noisy & 36.9 & 16.03 & 79.39 & 47.72 & 105.02 & 70.03 & 18.69 & 12.82 & 50.48 & 41.36 & 72.49 & 62.03 \\
PCN & 36.86 & 15.99 & 79.26 & 47.59 & 104.86 & 69.87 & 11.03 & 6.46 & 19.78 & 13.7 & 32.03 & 24.86 \\
GPDNet & 23.1 & 7.14 & 42.84 & 18.55 & 58.37 & 30.66 & 10.49 & 6.35 & 32.88 & 25.03 & 50.85 & 41.34 \\
DMRDenoise & 47.12 & 21.96 & 50.85 & 25.23 & 52.77 & 26.69 & 12.05 & 7.62 & 14.43 & 9.7 & 16.96 & 11.9 \\
PDFlow & 21.26 & 6.74 & 32.46 & 13.24 & 36.27 & 17.02 & 6.51 & 4.16 & 12.7 & 9.21 & 18.74 & 14.26 \\
ScoreDenoise & 25.22 & 7.54 & 36.83 & 13.8 & 42.32 & 19.04 & 7.16 & 4.0 & 12.89 & 8.33 & 14.45 & 9.58 \\
Pointfilter & 24.61 & 7.3 & 35.34 & 11.55 & 40.99 & 15.05 & 7.58 & 4.32 & 9.07 & 5.07 & 10.99 & 6.29 \\
\midrule
\textbf{Ours} & \textbf{20.56} & \textbf{5.01} & \textbf{30.43} & \textbf{8.45} & \textbf{33.52} & \textbf{10.45} & \textbf{6.05} & \textbf{3.02} & \textbf{8.03} & \textbf{4.36} & \textbf{10.15} & \textbf{5.88} \\
\bottomrule
\end{tabular}
\caption{Filtering results on the PUNet dataset. CD and P2M distances are multiplied by $10^5$.}
\label{tab:syn-data} 
\end{table*}
\midsepdefault

\begin{figure*}[!tp]
\centering
\includegraphics[width=0.93\linewidth]{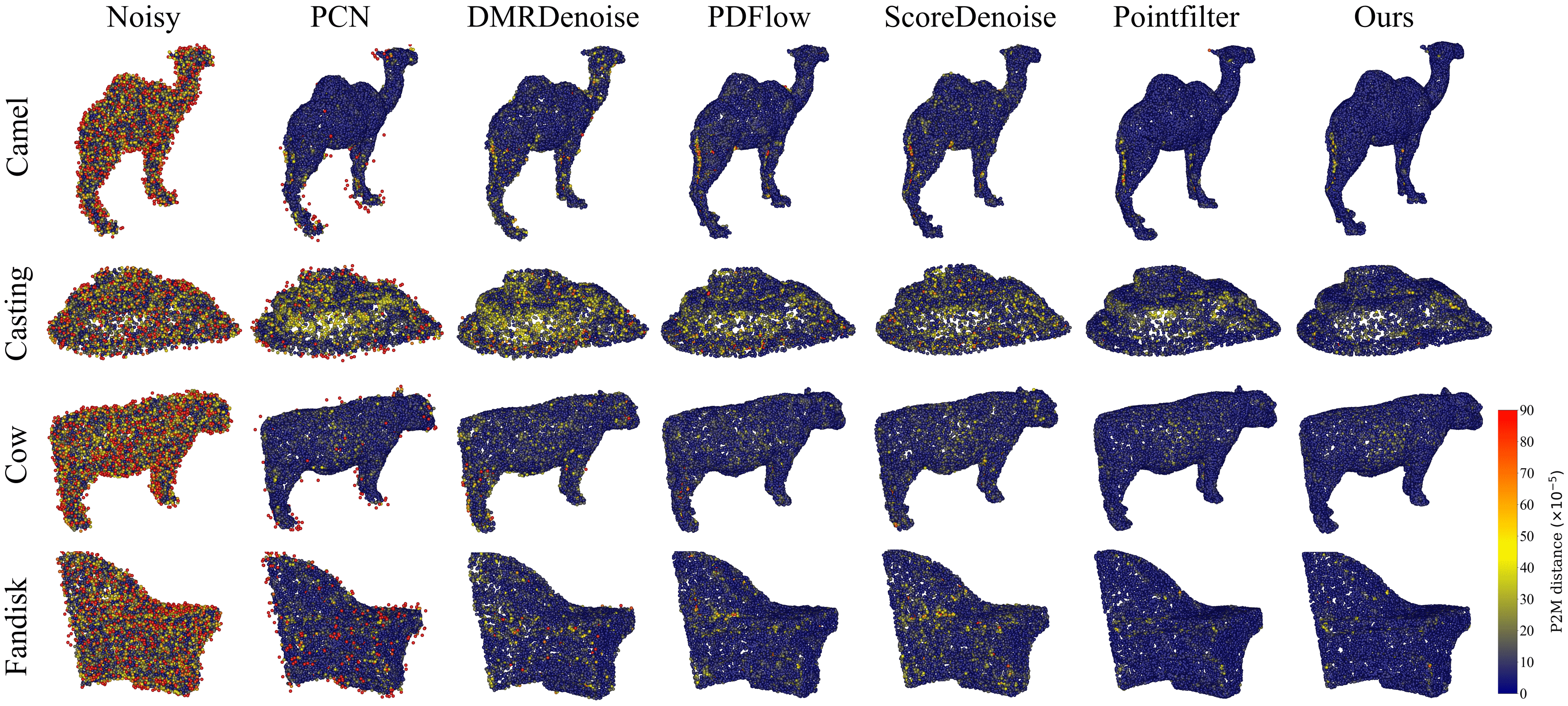}
\caption{Visual results of of point-wise P2M distance for 50K resolution shapes with Gaussian noise of 2\% of the bounding sphere radius. }
\label{fig:all-synthetic-results}
\end{figure*}

\section{Experimental Results}
\subsection{Dataset and implementation}
\label{subsec:dataset}
We follow ScoreDenoise~\cite{Luo-Score-Based-Denoising} and PDFlow~\cite{Mao-PDFlow} and utilize the PUNet~\cite{Yu-PUNet} dataset to train our network. %For comparison, we use the ScoreDenoise and PDFlow pretrained models, trained on the PUNet dataset, and 
\dt{We retrain all other methods including PCN~\cite{Rakotosaona-PCN}, GPDNet~\cite{Pistilli-GPDNet}, DMRDenoise~\cite{Luo-DMRDenoise} and Pointfilter~\cite{Zhang-Pointfilter}.} An implementation of RePCDNet~\cite{Chen-RePCD} was not available for comparison. Poisson disk sampling is used to sample 40 training meshes, at resolutions of 10K, 30K and 50K points, which yields 120 point clouds for the training set. Subsequently, Gaussian noise with standard deviation ranging from 0.5\% to 2\% of the bounding sphere's radius is added to each point cloud. For testing, we utilize 20 test meshes sampled at resolutions of 10K and 50K, similar to~\cite{Luo-Score-Based-Denoising, Mao-PDFlow}. These 40 point clouds are perturbed by Gaussian noise with standard deviations of 1\%, 2\% and 2.5\% of the bounding sphere's radius. For comparisons on real world scans, we look at test results on the following datasets: The Paris-Rue-Madame database consisting of scans acquired by the Mobile Laser Scanning system L3D2~\cite{Serna-Rue-Madame}, which capture the street of Rue Madame in the 6th district of Paris. It contains real world noisy artifacts, a consequence of the limitations of scanning technology, and provides an excellent basis for comparing performance on real-world data. As ground truth point clouds and meshes are unavailable, we present qualitative results. Next, we consider the Kinect v1 and Kinect v2 datasets of~\cite{Wang-Kinect} consisting of 71 and 72 real-world scans acquired using Microsoft Kinect v1 and Kinect v2 cameras.
% \textbf{Evaluation metrics}. 
\dt{We compare performance of all methods on the Chamfer distance (CD)~\cite{Fan-Chamfer} and the Point2Mesh distance (P2M)~\cite{Li-Dis-PU} evaluation metrics.} Both metrics are calculated using their latest implementations in PyTorch3D~\cite{Ravi-PyTorch3D}.

\textbf{Implementation}. Our IterativePFN network is trained on NVIDIA A100 GPUs using PyTorch 1.11.0 with CUDA 11.3. We train the network for 100 epochs, with the Adam optimizer and a learning rate of $1\times 10^{-4}$. All methods are tested on a NVIDIA GeForce RTX 3090 GPU to ensure fair comparison of test times.

\subsection{Results on synthetic data}
\label{subsec:res-syn-data}
The comparison of methods on synthetic data is given by Table~\ref{tab:syn-data} and Fig.~\ref{fig:all-synthetic-results}. The baseline displacement based methods, PCN and GPDNet, show sub-optimal performance at both low and high resolutions. Specifically, PCN shows sensitivity to high noise and filtered point clouds suffer from shrinkage. The resampling based DMRDenoise and score-matching based ScoreDenoise both show sensitivity to noise. The normalizing flow based PDFlow performs well at low resolution but has trouble at high resolution and high noise. Our method shows an advantage across all resolutions and noise scales. Pointfilter is also able to generalize well across resolutions and noise scales as it is based on a weighted bilateral filtering inspired loss function that uses ground truth normal information. However, it still compares less favorably to our method. Furthermore, for complex shapes such as the Casting shape of Fig.~\ref{fig:all-synthetic-results}, Pointfilter's weighted loss causes the collapse of closely neighboring surfaces onto a single surface.

\subsection{Results on scanned data}
\label{subsec:res-scan-data} 

\begin{figure*}[!tp]
\centering
\includegraphics[width=0.93\linewidth]{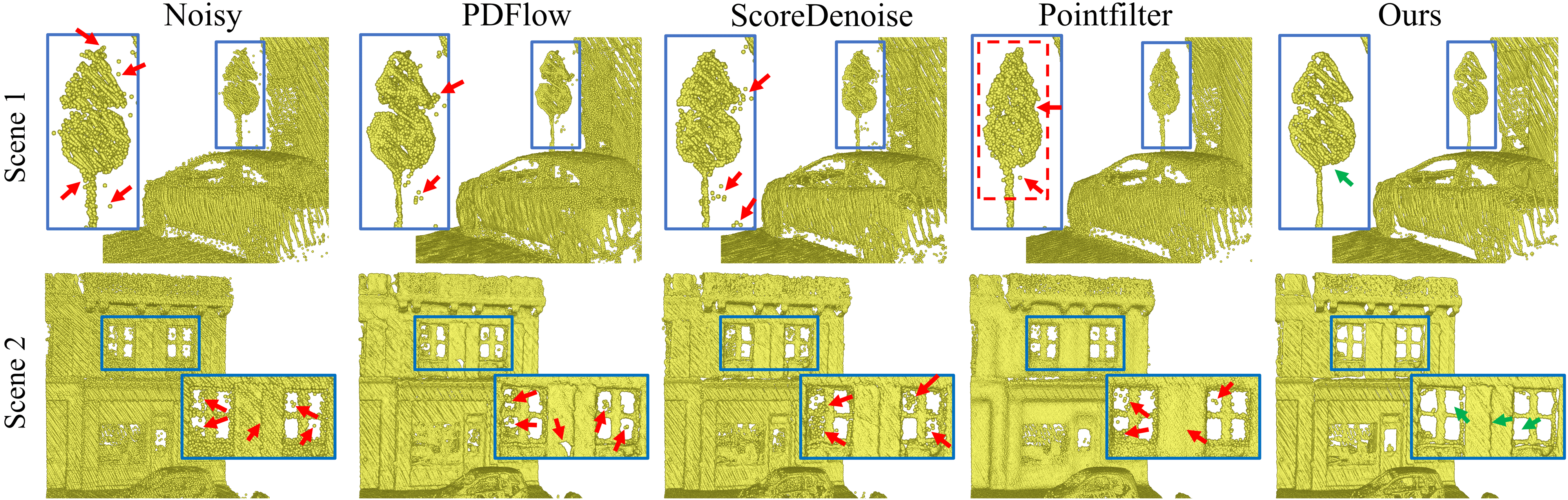}
\caption{\dt{Results on two scenes of the real-world RueMadame dataset. Green and red arrows are used to indicate accurately and inaccurately filtered regions, respectively. PDFlow and ScoreDenoise leave behind large outliers while Pointfilter distorts underlying shapes.}}
\label{fig:rm-results}
\end{figure*}

Fig.~\ref{fig:rm-results} demonstrates filtering results on two scenes of the RueMadame database. As it contains only noisy scanned data, we only consider visual results. As shown by the results on scene 1, only our method consistently filters surfaces such as the sign post and vehicle hood. PDFlow leaves behind many noisy artifacts while ScoreDenoise performs only a little better. Likewise, Pointfilter leaves noisy artifacts while taking a large amount of time to finish filtering the scene. This is due to their filtering strategy that takes \textbf{o}ne \textbf{p}atch \textbf{p}er central \textbf{p}oint (OPPP). Therefore, patches need to be constructed for each point, which is inefficient. Please refer to the supplementary document for details on each method's runtime. Table~\ref{tab:scan-data} presents results on the Kinect v1 and v2 datasets. Our method, with a smaller patch size of 100, with edges between 32 $k$-nearest neighbors, achieves the best results on the CD metric and second best results on the P2M metric. Additional results on synthetic and scanned data is provided in the supplementary.

\subsection{Ablation study}
\label{sec:ablations}
In this section we look at the importance of the following elements to our method:
1) The number of IterationModules modelling the iteration number in our IterativePFN network, \dt{2) the impact of true iterative filtering and 3)} the impact of a fixed ground truth target, given by Eq.~\eqref{eq:fixed-gt-nn-loss}, versus an adaptive ground truth target, Eq.~\eqref{eq:initial-nn-loss} and Eq.~\eqref{eq:initial-nn-loss-full}. \dt{The fixed ground truth target-based loss is given by,}

\dt{
\begin{align}
\label{eq:fixed-gt-nn-loss}
\mathcal{L}_b = \sum^{T}_{\tau=1}\left[\sum_i w_i \left(\norm{\pmb{d}^{(\tau)}_i - \pmb{\delta x}^{(\tau)}_i(\mathcal{Y})}^2_2\right)\right],
\end{align}
where $\pmb{\delta x}^{(\tau)}_i(\mathcal{Y})=NN(\pmb{x}^{(\tau-1)}_i, \mathcal{Y})-\pmb{x}^{(\tau-1)}_i$.} Table~\ref{tab:ablation} demonstrates the impact of iteration number on filtering. Although 8 iterations, i.e., 8 IterationModules, provide the best results for 1\% and 2\% noise, it does not generalize well, compared to 4 iterations, on the unseen noise of 2.5\%. This is due to the higher number of IterationModules causing the network to specialize on the training noise scales of 0.5\% to 2\%. Furthermore, 8 IterationModules lead to \dt{larger network size and memory usage. Runtime memory consumption for 1, 2, 4, 8 and 12 iterations are 3.9GB, 7.6GB, 15GB, 29.5GB and 44.1GB, respectively.} Hence, we select 4 iterations as the optimal number. \dt{Moreover, to confirm the efficacy of true iterative filtering, we train a deep network of 1 ItM (DPFN), with 6 Dynamic EdgeConv layers, and the same number of parameters (3.2M) as our IterativePFN with 4 ItMs. This variant performs sub-optimally, especially at high noise. This indicates the importance of the true iterative filtering approach.} Finally, we consider the impact of keeping the ground truth target fixed during training, i.e., $\mathcal{L}_b$, where $\mathcal{Y}$ corresponds to the clean patch. We see that the adaptive ground truth $\mathcal{L}_a$ makes a noticeable difference, especially at the unseen noise scale of 2.5\%. \dt{The supplementary contains additional ablation studies on 1) the impact of iteration number at higher noise, 2) the effect of patch stitching on filtering and 3) filtering results when using PointNet++ as the encoder backbone within each ItM.}

\midsepremove
\begin{table}[!tp]
\centering
\begin{tabular}{l|ll|ll}
\toprule
\multicolumn{1}{c|}{\multirow{2}{*}{Method}} & \multicolumn{2}{c|}{Kinect v1} & \multicolumn{2}{c}{Kinect v2} \\
\cmidrule{2-5}
 & CD & P2M & CD & P2M \\
\midrule
Noisy & 14.49 & 9.32 & 22.63 & 13.39 \\
PCN & 13.73 & 8.75 & 22.48 & 13.29 \\
GPDNet & 14.83 & 8.69 & 23.09 & 11.78 \\
DMRDenoise & 22.78 & 12.89 & \multicolumn{1}{c}{-} & \multicolumn{1}{c}{-} \\
PDFlow & 13.34 & 8.69 & 19.87 & 11.69 \\
ScoreDenoise & 13.22 & {\ul 8.18} & 19.66 & 11.08 \\
Pointfilter & 13.77 & \textbf{7.91} & {\ul 18.85} & \textbf{10.29} \\
\midrule
Ours ($|\mathcal{X}|=1000$) & 14.07 & 9.16 & 19.68 & 11.69 \\
\textbf{Ours} ($|\mathcal{X}|=100$) & \textbf{13.2} & 8.43 & \textbf{18.69} & {\ul 10.92} \\
\bottomrule
\end{tabular}
\caption{\dt{Results on the Kinect v1 and Kinect v2 datasets. CD and P2M distances are multiplied by $10^5$.} }
\label{tab:scan-data} 
\end{table}
\midsepdefault

\midsepremove
\begin{table}[!tp]
\centering
\setlength\tabcolsep{4pt} % default value: 6pt
\begin{tabular}{c|ll|ll|ll}
\toprule
\multirow{3}{*}{Ablation} & \multicolumn{6}{c}{10K points} \\
\cmidrule{2-7}
& \multicolumn{2}{c|}{1\% noise} & \multicolumn{2}{c|}{2\% noise} & \multicolumn{2}{c}{2.5\% noise} \\
\cmidrule{2-7}
& CD & P2M & CD & P2M & CD & P2M \\ 
\midrule
$\mathcal{L}_a$ \& 1 it. & 21.95 & 5.42 & 32.38 & 9.55 & 36.98 & 12.71 \\
$\mathcal{L}_a$ \& 2 it. & 21.13 & 5.14 & 30.82 & 8.67 & 34.33 & 11.0 \\
$\mathcal{L}_a$ \& 4 it. & \uline{20.56} & \uline{5.01} & \uline{30.43} & \uline{8.45} & \textbf{33.52} & \textbf{10.45} \\
$\mathcal{L}_a$ \& 8 it. & \textbf{19.78} & \textbf{4.9} & \textbf{30.12} & \textbf{8.3} & \uline{33.88} & \uline{10.78} \\
$\mathcal{L}_a$ \& 12 it. & 20.49 & 5.23 & 30.64 & 8.87 & 34.46 & 11.25 \\
$\mathcal{L}_a$ \& DPFN & 21.03 & 5.05 & 30.96 & 8.53 & 35.2 & 11.4  \\
\midrule
$\mathcal{L}_b$ \& 4 it. & 20.64 & 5.04 & 30.59 & 8.54 & 34.17 & 10.87 \\
\bottomrule
\end{tabular}
\caption{Ablation results for different iteration numbers and different loss functions. CD and P2M distances are multiplied by $10^5$.}
\label{tab:ablation} 
\end{table}
\midsepdefault

\section{Limitations and Future Work} The adaptive ground truth targets rely on adding noise, of a given distribution and noise scale (i.e., standard deviation), during training. However, this implies the noise distribution should be easy to replicate, such as Gaussian noise, with different scales in decreasing order. In future, it would be interesting to generalize this approach to utilize noisy data that simulates real world noise, with noise distributions similar to that of real world scanners, such as Lidar scanner data. Additionally, we hope to apply our novel network architecture, of stacked encoder-decoder pairs, to other tasks, e.g., point cloud upsampling, that may benefit from a true iterative approach.

\section{Conclusion}
In this paper, we present IterativePFN, which consists of multiple IterationModules that explicitly model the iterative filtering process internally, unlike state-of-the-art learning based methods which only perform iterative filtering at test time. Furthermore, state-of-the-art methods attempt to learn filtered displacements that directly shift noisy points to the underlying clean surfaces and neglect the relationship between intermediate filtered points. Our IterativePFN network employs a re-imagined loss function that utilizes an adaptive ground truth target at each iteration to capture this relationship between intermediate filtered results during training. Our method ensures filtered results converge to the clean surface faster and, overall, performs better than state-of-the-art methods. 

%%%%%%%%% REFERENCES
{\small
\bibliographystyle{ieee_fullname}
\bibliography{egbib}

\begin{thebibliography}{10}\itemsep=-1pt

\bibitem{IMLS-Adamson}
A. Adamson and M. Alexa.
\newblock Point-sampled cell complexes.
\newblock {\em ACM Trans. Graph.}, 25:671--680, 2006.

\bibitem{Alexa--MLS-PSS}
M. Alexa, J. Behr, D. Cohen-Or, S. Fleishman, D. Levin, and Cláudio~T. Silva.
\newblock Computing and rendering point set surfaces.
\newblock {\em IEEE Trans. Vis. Comput. Graph.}, 9:3--15, 2003.

\bibitem{Avron-L1}
H. Avron, Andrei Sharf, C. Greif, and D. Cohen-Or.
\newblock L1-sparse reconstruction of sharp point set surfaces.
\newblock {\em ACM Trans. Graph.}, 29:135:1--135:12, 2010.

\bibitem{Ben-Shabat-DeepFit}
Yizhak Ben-Shabat and Stephen Gould.
\newblock Deepfit: 3d surface fitting via neural network weighted least
  squares.
\newblock In {\em Computer Vision – ECCV 2020}, pages 20--34. Springer
  International Publishing, 2020.

\bibitem{Cai-ShapeGF}
Ruojin Cai, Guandao Yang, Hadar Averbuch-Elor, Zekun Hao, Serge Belongie, Noah
  Snavely, and Bharath Hariharan.
\newblock Learning gradient fields for shape generation.
\newblock In {\em Computer Vision – ECCV 2020}, pages 364--381. Springer
  International Publishing, 2020.

\bibitem{Cazals-Jet}
Frederic Cazals and Marc Pouget.
\newblock Estimating differential quantities using polynomial fitting of
  osculating jets.
\newblock {\em Computer Aided Geometric Design}, 22(2):121--146, 2005.

\bibitem{Chen-RePCD}
Honghua Chen, Zeyong Wei, Xianzhi Li, Yabin Xu, Mingqiang Wei, and Jun Wang.
\newblock Repcd-net: Feature-aware recurrent point cloud denoising network.
\newblock {\em International Journal of Computer Vision}, 130(3):615--629,
  2022.

\bibitem{Digne-Similarity}
Julie Digne.
\newblock Similarity based filtering of point clouds.
\newblock {\em 2012 IEEE Computer Society Conference on Computer Vision and
  Pattern Recognition Workshops}, pages 73--79, 2012.

\bibitem{Digne-Bilateral}
Julie Digne and C.~D. Franchis.
\newblock The bilateral filter for point clouds.
\newblock {\em Image Process. Line}, 7:278--287, 2017.

\bibitem{Fan-Chamfer}
Haoqiang Fan, Hao Su, and Leonidas~J. Guibas.
\newblock A point set generation network for 3d object reconstruction from a
  single image.
\newblock In {\em Proceedings of the IEEE Conference on Computer Vision and
  Pattern Recognition (CVPR)}, 2017.

\bibitem{Fleishman-Bilateral}
S. Fleishman, Iddo Drori, and D. Cohen-Or.
\newblock Bilateral mesh denoising.
\newblock {\em ACM SIGGRAPH 2003 Papers}, 2003.

\bibitem{APSS-Guennebaud}
Gaël Guennebaud and M. Gross.
\newblock Algebraic point set surfaces.
\newblock In {\em SIGGRAPH 2007}, 2007.

\bibitem{Hoppe-PCA}
Hugues Hoppe, T. DeRose, T. Duchamp, J. McDonald, and W. Stuetzle.
\newblock Surface reconstruction from unorganized points.
\newblock {\em Proceedings of the 19th annual conference on Computer graphics
  and interactive techniques}, 1992.

\bibitem{Huang-WLOP}
Hui Huang, Dan Li, Hongxing Zhang, U. Ascher, and D. Cohen-Or.
\newblock Consolidation of unorganized point clouds for surface reconstruction.
\newblock {\em ACM SIGGRAPH Asia 2009 papers}, 2009.

\bibitem{Huang-EAR}
Hui Huang, Shihao Wu, Minglun Gong, D. Cohen-Or, U. Ascher, and Hongxing Zhang.
\newblock Edge-aware point set resampling.
\newblock {\em ACM Transactions on Graphics (TOG)}, 32:1 -- 12, 2013.

\bibitem{Kim-3D-Printing}
Youngki Kim, Kiyoun Kwon, and Duhwan Mun.
\newblock Mesh-offset-based method to generate a delta volume to support the
  maintenance of partially damaged parts through 3d printing.
\newblock {\em Journal of Mechanical Science and Technology}, 35(7):3131--3143,
  2021.

\bibitem{MLS-Levin}
D. Levin.
\newblock The approximation power of moving least-squares.
\newblock {\em Math. Comput.}, 67:1517--1531, 1998.

\bibitem{Li-Dis-PU}
Ruihui Li, Xianzhi Li, Pheng-Ann Heng, and Chi-Wing Fu.
\newblock Point cloud upsampling via disentangled refinement.
\newblock In {\em Proceedings of the IEEE Conference on Computer Vision and
  Pattern Recognition (CVPR)}, 2021.

\bibitem{Lipman-LOP}
Y. Lipman, D. Cohen-Or, D. Levin, and H. Tal-Ezer.
\newblock Parameterization-free projection for geometry reconstruction.
\newblock {\em ACM SIGGRAPH 2007 papers}, 2007.

\bibitem{Lu-Low-Rank}
Xuequan Lu, S. Schaefer, Jun Luo, Lizhuang Ma, and Y. He.
\newblock Low rank matrix approximation for 3d geometry filtering.
\newblock {\em IEEE transactions on visualization and computer graphics}, PP,
  2020.

\bibitem{Luo-Pillar-Motion}
Chenxu Luo, Xiaodong Yang, and A. Yuille.
\newblock Self-supervised pillar motion learning for autonomous driving.
\newblock In {\em CVPR}, 2021.

\bibitem{Luo-DMRDenoise}
Shitong Luo and Wei Hu.
\newblock Differentiable manifold reconstruction for point cloud denoising.
\newblock In {\em Proceedings of the 28th ACM International Conference on
  Multimedia}, page 1330–1338. Association for Computing Machinery, 2020.

\bibitem{Luo-Diffusive}
Shitong Luo and Wei Hu.
\newblock Diffusion probabilistic models for 3d point cloud generation.
\newblock In {\em Proceedings of the IEEE/CVF Conference on Computer Vision and
  Pattern Recognition (CVPR)}, pages 2837--2845, 2021.

\bibitem{Luo-Score-Based-Denoising}
Shitong Luo and Wei Hu.
\newblock Score-based point cloud denoising.
\newblock In {\em Proceedings of the IEEE/CVF International Conference on
  Computer Vision (ICCV)}, pages 4583--4592, October 2021.

\bibitem{Mao-PDFlow}
Aihua Mao, Zihui Du, Yu-Hui Wen, Jun Xuan, and Yong-Jin Liu.
\newblock Pd-flow: A point cloud denoising framework with normalizing flows.
\newblock In {\em The European Conference on Computer Vision (ECCV)}, 2022.

\bibitem{Pistilli-GPDNet}
Francesca Pistilli, Giulia Fracastoro, Diego Valsesia, and Enrico Magli.
\newblock Learning graph-convolutional representations for point cloud
  denoising.
\newblock In {\em Computer Vision – ECCV 2020}, pages 103--118. Springer
  International Publishing, 2020.

\bibitem{Preiner-CLOP}
R. Preiner, O. Mattausch, Murat Arikan, R. Pajarola, and M. Wimmer.
\newblock Continuous projection for fast l1 reconstruction.
\newblock {\em ACM Transactions on Graphics (TOG)}, 33:1 -- 13, 2014.

\bibitem{Qi-PointNet}
C. Qi, Hao Su, Kaichun Mo, and L. Guibas.
\newblock Pointnet: Deep learning on point sets for 3d classification and
  segmentation.
\newblock {\em 2017 IEEE Conference on Computer Vision and Pattern Recognition
  (CVPR)}, pages 77--85, 2017.

\bibitem{Qi-PointNet++}
Charles~Ruizhongtai Qi, Li Yi, Hao Su, and Leonidas~J. Guibas.
\newblock Pointnet++: Deep hierarchical feature learning on point sets in a
  metric space.
\newblock In {\em Advances in Neural Information Processing Systems},
  volume~30. Curran Associates, Inc., 2017.

\bibitem{Rakotosaona-PCN}
Marie-Julie Rakotosaona, Vittorio~La Barbera, Paul Guerrero, N. Mitra, and M.
  Ovsjanikov.
\newblock Pointcleannet: Learning to denoise and remove outliers from dense
  point clouds.
\newblock {\em Computer Graphics Forum}, 39, 2020.

\bibitem{Ravi-PyTorch3D}
Nikhila Ravi, Jeremy Reizenstein, David Novotny, Taylor Gordon, Wan-Yen Lo,
  Justin Johnson, and Georgia Gkioxari.
\newblock Accelerating 3d deep learning with pytorch3d.
\newblock {\em ArXiv}, 2020.

\bibitem{Remil-Data-Driven-Sparse-Priors}
Oussama Remil, Qian Xie, Xingyu Xie, Kai Xu, and J. Wang.
\newblock Data driven sparse priors of 3d shapes.
\newblock {\em Computer Graphics Forum}, 36, 2017.

\bibitem{Roveri-PointProNets}
Riccardo Roveri, A.~Cengiz Öztireli, Ioana Pandele, and Markus Gross.
\newblock Pointpronets: Consolidation of point clouds with convolutional neural
  networks.
\newblock {\em Computer Graphics Forum}, 37(2):87--99, 2018.

\bibitem{Serna-Rue-Madame}
Andr{\'e}s Serna, Beatriz Marcotegui, François Goulette, and Jean-Emmanuel
  Deschaud.
\newblock Paris-rue-madame database - a 3d mobile laser scanner dataset for
  benchmarking urban detection, segmentation and classification methods.
\newblock In {\em ICPRAM}, 2014.

\bibitem{Song-Score-Matching}
Yang Song and Stefano Ermon.
\newblock Generative modeling by estimating gradients of the data distribution.
\newblock In {\em Proceedings of the 33rd International Conference on Neural
  Information Processing Systems}. Curran Associates Inc., 2019.

\bibitem{Sun-L0}
Yujing Sun, S. Schaefer, and Wenping Wang.
\newblock Denoising point sets via l0 minimization.
\newblock {\em Comput. Aided Geom. Des.}, 35-36:2--15, 2015.

\bibitem{Urech-Urban-Planning}
Philipp R.~W. Urech, M. Dissegna, C. Girot, and A. Grêt-Regamey.
\newblock Point cloud modeling as a bridge between landscape design and
  planning.
\newblock {\em Landscape and Urban Planning}, 203:103903, 2020.

\bibitem{Wang-Kinect}
Peng-Shuai Wang, Yang Liu, and Xin Tong.
\newblock Mesh denoising via cascaded normal regression.
\newblock {\em ACM Trans. Graph.}, 35(6):Article 232, 2016.

\bibitem{Wang-DGCNN}
Yue Wang, Yongbin Sun, Ziwei Liu, Sanjay~E. Sarma, Michael~M. Bronstein, and
  Justin~M. Solomon.
\newblock Dynamic graph cnn for learning on point clouds.
\newblock {\em ACM Transactions on Graphics (TOG)}, 2019.

\bibitem{Wen-PMPNet}
X Wen, P Xiang, Z Han, Y Cao, P Wan, W Zheng, and Y Liu.
\newblock Pmp-net: Point cloud completion by learning multi-step point moving
  paths.
\newblock In {\em 2021 IEEE/CVF Conference on Computer Vision and Pattern
  Recognition (CVPR)}, pages 7439--7448, 2021.

\bibitem{Bekiroglu-PCD-Robotics}
Bekiroglu Yasemin, Björkman Mårten, Gandler Gabriela~Zarzar, Exner Johannes,
  Ek Carl~Henrik, and Kragic Danica.
\newblock Visual and tactile 3d point cloud data from real robots for shape
  modeling and completion.
\newblock {\em Data in Brief}, 30(105335-), 2020.

\bibitem{Yu-PUNet}
Lequan Yu, Xianzhi Li, Chi-Wing Fu, Daniel Cohen-Or, and Pheng-Ann Heng.
\newblock Pu-net: Point cloud upsampling network.
\newblock In {\em Proceedings of the IEEE Conference on Computer Vision and
  Pattern Recognition (CVPR)}, 2018.

\bibitem{Yu-EC-Net}
L. Yu, X. Li, C.~W. Fu, P.~A. Heng, and D. Cohen-Or.
\newblock {\em EC-Net: An edge-aware point set consolidation network}, volume
  11211 LNCS of {\em Lecture Notes in Computer Science}.
\newblock Springer Verlag, 2018.

\bibitem{Zhang-Pointfilter}
Dongbo Zhang, Xuequan Lu, Hong Qin, and Y. He.
\newblock Pointfilter: Point cloud filtering via encoder-decoder modeling.
\newblock {\em IEEE Transactions on Visualization and Computer Graphics},
  27:2015--2027, 2021.

\bibitem{Zhou-Patch-Stitching}
Jun Zhou, Wei Jin, Mingjie Wang, Xiuping Liu, Zhiyang Li, and Zhaobin Liu.
\newblock Fast and accurate normal estimation for point clouds via patch
  stitching.
\newblock {\em Computer-Aided Design}, 142, 2022.

\bibitem{Zhu-AdaFit}
Runsong Zhu, Yuan Liu, Zhen Dong, Yuan Wang, Tengping Jiang, Wenping Wang, and
  Bisheng Yang.
\newblock Adafit: Rethinking learning-based normal estimation on point clouds.
\newblock In {\em Proceedings of the IEEE/CVF International Conference on
  Computer Vision (ICCV)}, pages 6118--6127, October 2021.

\end{thebibliography}
}

\setcounter{section}{0}
\renewcommand\thesection{\Alph{section}}
\renewcommand\thesubsection{\thesection.\arabic{subsection}}

\title{Supplementary Material of IterativePFN: True Iterative Point Cloud Filtering }

\author{Dasith de Silva Edirimuni$^1$,~Xuequan Lu$^1$,~Zhiwen Shao$^2$,~Gang Li$^1$,~Antonio Robles-Kelly$^{1,4}$,~Ying He$^3$ \\
$^1$School of Information Technology, Deakin University\\
$^2$School of Computer Science and Technology, China University of Mining and Technology \\
$^3$School of Computer Science and Engineering, Nanyang Technological University \\
$^4$Defense Science and Technology Group, Australia \\
\tt\small{\{dtdesilva, xuequan.lu, gang.li, antonio.robles-kelly\}@deakin.edu.au,}
\\
\tt\small{zhiwen\_shao@cumt.edu.cn, yhe@ntu.edu.sg}
}

\maketitle

%%%%%%%%% BODY TEXT
In this supplementary document, to the main paper, we provide the following:
\begin{enumerate}[label*=\Alph*.]
\item Further comparisons on synthetic and scanned data
\begin{enumerate}[label*=\arabic*.]
\item Additional visual comparisons on the Rue Madame and Kinect datasets 
\item Comparison of conventional methods on the PUNet dataset with Gaussian noise and Kinect dataset
\item Quantitative comparisons on PUNet dataset with different noise patterns
\end{enumerate}
\item \dt{Further ablations on iteration number }
\item Ablation study on patch stitching
\item \dt{Filtering results for PointNet++ based encoders}
\item Runtimes for learning based methods
% \item The implementation code has been made available at an \textit{anonymous} github page:  \url{https://github.com/ml-anon-user/tags-anon-repo}.
\end{enumerate}

\section{Further Comparisons on Synthetic and Scanned Data}
In this section, we provide additional visual and quantitative results on synthetic and scanned data. 
\subsection{Additional visual comparisons on the Rue Madame and Kinect datasets}
\begin{figure*}[!tp]
\centering
\includegraphics[width=0.97\linewidth]{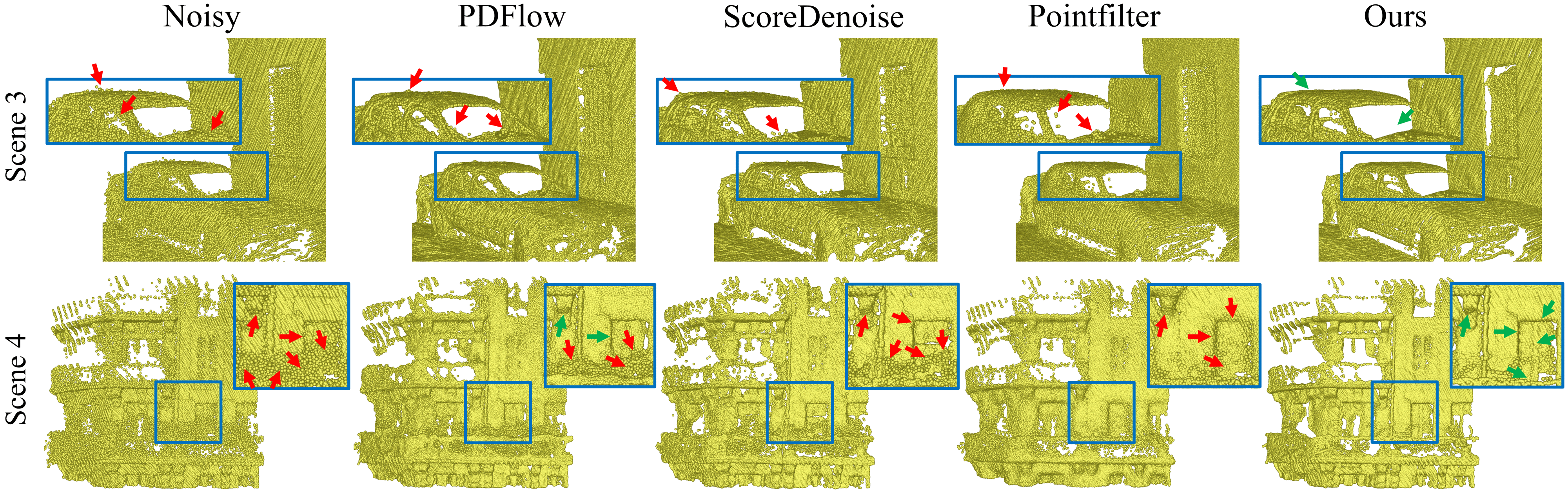}
\caption{\dt{Visual results on two additional scenes of the RueMadame dataset. Green and red arrows are used to indicate accurately and inaccurately
filtered regions, respectively.}}
\label{fig:supp-rm-results}
\end{figure*}

\begin{figure*}[!tp]
\centering
\includegraphics[width=0.97\linewidth]{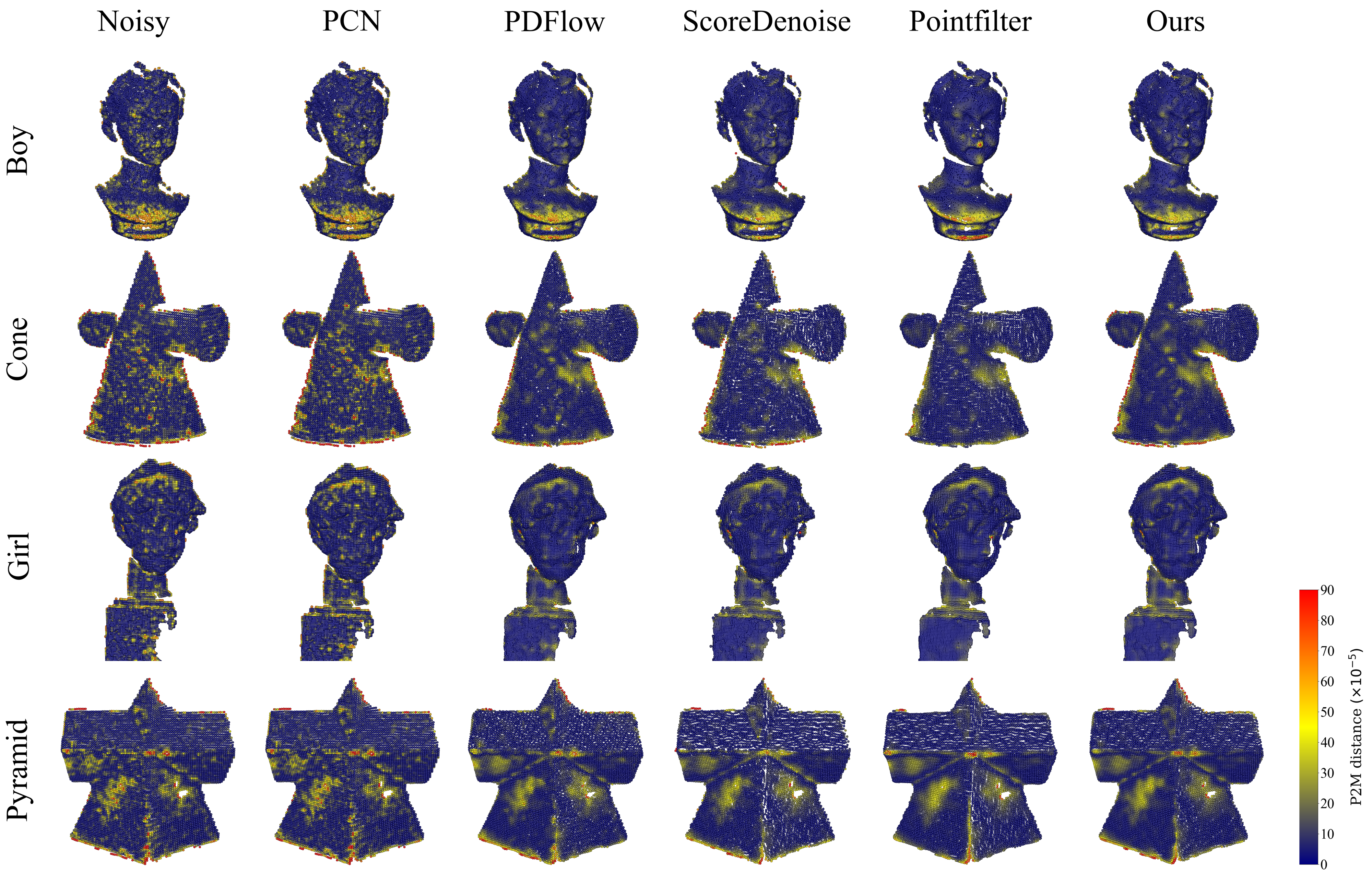}
\caption{Visual results of point-wise P2M distances for 4 scans of the Kinect v2 dataset.}
\label{fig:supp-kinect-results}
\end{figure*}

Table~\ref{tab:scan-data} and Fig.~\ref{fig:rm-results}, of the main paper, present quantitative results on the Kinect dataset and visual results on the RueMadame dataset, respectively. To further demonstrate our method's robustness, in the presence of real world noise, we provide visual results on two additional scenes of the RueMadame dataset, given in Fig.~\ref{fig:supp-rm-results}. Moreover, we provide visual comparisons of 4 scans from the Kinect v2 dataset in Fig.~\ref{fig:supp-kinect-results}, which could not be included in the main paper due to constraints of space.

As seen in Fig.~\ref{fig:supp-rm-results}, our method performs considerably better compared to other methods when filtering noise for the RueMadame scenes. In scene 3, the roof and body of the vehicle is best filtered by our method while the filtering results of other methods leave behind significant noise artifacts. Furthermore, as illustrated in Fig.~\ref{fig:supp-kinect-results}, we are able to better filter complex Kinect v2 scans, such as the model of the Boy, as opposed to state-of-the-art methods. This is supported by the results presented in Table~\ref{tab:scan-data} of the main paper, where we achieve the best results on the CD metric. It indicates that our method produces filtered point clouds closer in point distribution, and proximity, to the clean point clouds of the Kinect v2 dataset, than other state-of-the-art methods.

\subsection{Comparison of conventional methods on the PUNet dataset with Gaussian noise and Kinect dataset}
In the main paper we were not able to present a comparison of the filtering performance of conventional methods, due to limited space. In Table~\ref{tab:syn-data-conventional} and Table~\ref{tab:scan-data-conventional}, we present quantitative results for the Bilateral filter~\cite{Digne-Bilateral}, Jet fitting mechanism~\cite{Cazals-Jet} and WLOP regularization mechanism~\cite{Huang-WLOP}. The Bilater filter relies on initial point normals which were calculated using Principal Component Analysis (PCA)~\cite{Hoppe-PCA} as it is a widely used normal estimation technique. WLOP regularizes and downsamples noisy point clouds. These must then be upsampled and we use the Edge Aware Resampling (EAR) mechanism of Huang et al.~\cite{Huang-EAR} to accomplish this.

In general, conventional methods perform poorly compared to learning based methods. In addition, they require parameter tuning to obtain best results which can be tedious and time-consuming. Methods such as the Bilateral filter further rely on initial point normals to filter effectively and the accuracy of these normals have an effect on the overall filtered output. WLOP, by contrast, recovers good P2M results on the Kinect v1 dataset. This is possibly due to the resampling procedure where the downsampled output from WLOP is upsampled by the EAR mechanism. Although this is reasonably successful for the Kinect v1 scanned data with relatively simple topology, for more complex topologies as those associated with the shapes in the PUNet dataset, the resampling procedure is less successful. As shown in Table~\ref{tab:syn-data-conventional}, the resampling procedure yields high CD and P2M values as the WLOP mechanism is not able to optimally identify clean surfaces and omits geometric details. Moreover, WLOP performs poorly on the Kinect v2 dataset which illustrates its inability to generalize well across different filtering scenarios.

\midsepremove
\begin{table*}[!tp]
\centering
\begin{tabular}{l|llllll|llllll}
\toprule
\multicolumn{1}{c|}{\multirow{3}{*}{Method}} & \multicolumn{6}{c|}{10K points} & \multicolumn{6}{c}{50K points} \\
\cmidrule{2-13}
 & \multicolumn{2}{c|}{1\% noise} & \multicolumn{2}{c|}{2\% noise} & \multicolumn{2}{c|}{2.5\% noise} & \multicolumn{2}{c|}{1\% noise} & \multicolumn{2}{c|}{2\% noise} & \multicolumn{2}{c}{2.5\% noise} \\
\cmidrule{2-13}
 & CD & P2M & CD & P2M & CD & P2M & CD & P2M & CD & P2M & CD & P2M \\
\midrule
Bilateral & 31.47 & 14.95 & 51.94 & 26.75 & 73.83 & 44.31 & 17.09 & 13.9 & 17.98 & 13.38 & 25.34 & 19.55 \\
Jet & 31.91 & 13.42 & 55.25 & 31.14 & 61.51 & 36.56 & 7.95 & 4.39 & 13.67 & 8.77 & 16.68 & 11.18 \\
WLOP & 60.41 & 30.85 & 88.59 & 52.95 & 109.68 & 74.29 & 9.61 & 6.27 & 19.94 & 14.53 & 28.44 & 21.66 \\
\midrule
\textbf{Ours} & \textbf{20.56} & \textbf{5.01} & \textbf{30.43} & \textbf{8.45} & \textbf{33.52} & \textbf{10.45} & \textbf{6.05} & \textbf{3.02} & \textbf{8.03} & \textbf{4.36} & \textbf{10.15} & \textbf{5.88} \\
\bottomrule
\end{tabular}
\caption{Filtering results of conventional methods on the PUNet dataset with Gaussian noise. CD and P2M distances are multiplied by $10^5$.}
\label{tab:syn-data-conventional} 
\end{table*}
\midsepdefault

\midsepremove
\begin{table}[!ht]
\centering
\begin{tabular}{l|ll|ll}
\toprule
\multicolumn{1}{c|}{\multirow{2}{*}{Method}} & \multicolumn{2}{c|}{Kinect v1} & \multicolumn{2}{c}{Kinect v2} \\
\cmidrule{2-5}
 & CD & P2M & CD & P2M \\
\midrule
Bilateral & 13.65 & 9.27 & 20.14 & 12.1 \\
Jet & \uline{13.41} & 8.78 & \uline{19.82} & \uline{11.82} \\
WLOP & 13.89 & \textbf{7.65} & 33.0 & 14.79 \\
\midrule
\textbf{Ours} & \textbf{13.2} & \uline{8.43} & \textbf{18.69} & \textbf{10.92} \\
\bottomrule
\end{tabular}
\caption{Filtering results on the Kinect v1 and Kinect v2 datasets. CD and P2M distances are multiplied by $10^5$.}
\label{tab:scan-data-conventional} 
\end{table}
\midsepdefault

\subsection{Quantitative comparisons on PUNet dataset with different noise patterns}
\label{subsec:other-noise-patterns}
\midsepremove
\begin{table*}[!tp]
\centering
\begin{tabular}{l|llllll|llllll}
\toprule
\multicolumn{1}{c|}{\multirow{3}{*}{Method}} & \multicolumn{6}{c|}{10K points} & \multicolumn{6}{c}{50K points} \\
\cmidrule{2-13}
 & \multicolumn{2}{c|}{1\% noise} & \multicolumn{2}{c|}{2\% noise} & \multicolumn{2}{c|}{2.5\% noise} & \multicolumn{2}{c|}{1\% noise} & \multicolumn{2}{c|}{2\% noise} & \multicolumn{2}{c}{2.5\% noise} \\
\cmidrule{2-13}
 & CD & P2M & CD & P2M & CD & P2M & CD & P2M & CD & P2M & CD & P2M \\
\midrule
Noisy & 36.17 & 15.9 & 78.88 & 47.75 & 104.2 & 69.69 & 18.55 & 12.8 & 50.37 & 41.38 & 72.32 & 62.0 \\
\midrule
Bilateral & 30.92 & 14.98 & 52.56 & 27.77 & 74.19 & 45.08 & 17.08 & 13.99 & 18.99 & 14.52 & 27.72 & 22.07 \\
Jet & 31.57 & 13.54 & 55.39 & 31.59 & 60.89 & 35.89 & 7.87 & 4.38 & 13.84 & 9.0 & 17.21 & 11.74 \\
WLOP & 60.02 & 29.87 & 92.14 & 56.69 & 108.78 & 70.55 & 9.67 & 6.34 & 20.53 & 15.13 & 29.59 & 22.68 \\
\midrule
PCN & 36.12 & 15.86 & 78.74 & 47.6 & 104.04 & 69.53 & 10.82 & 6.37 & 20.49 & 14.42 & 34.17 & 26.98 \\
GPDNet & 22.71 & 7.17 & 43.34 & 19.24 & 59.25 & 31.68 & 10.54 & 6.46 & 33.09 & 25.38 & 51.0 & 41.67 \\
DMRDenoise & 47.8 & 22.76 & 50.96 & 25.26 & 53.47 & 27.62 & 12.05 & 7.6 & 14.63 & 9.9 & 17.57 & 12.48 \\
PDFlow & 21.03 & 6.78 & 32.93 & 13.76 & 37.12 & 17.81 & 6.53 & 4.19 & 13.12 & 9.6 & 20.56 & 15.97 \\
ScoreDenoise & 24.79 & 7.54 & 37.01 & 14.09 & 42.41 & 19.1 & 7.11 & 4.0 & 13.19 & 8.62 & 14.92 & 9.96 \\
Pointfilter & 23.99 & 7.2 & 35.29 & 11.84 & 41.57 & 15.71 & 7.57 & 4.36 & 9.64 & 5.6 & 12.4 & 7.51 \\
\midrule
\textbf{Ours} & \textbf{19.94} & \textbf{4.94} & \textbf{30.3} & \textbf{8.6} & \textbf{34.39} & \textbf{11.23} & \textbf{6.02} & \textbf{3.05} & \textbf{8.3} & \textbf{4.6} & \textbf{11.52} & \textbf{6.99} \\
\bottomrule
\end{tabular}
\caption{Filtering results on the PUNet dataset with non-isotropic Gaussian noise. CD and P2M distances are multiplied by $10^5$.}
\label{tab:syn-data-cov} 
\end{table*}
\midsepdefault

\begin{figure*}[!tp]
\centering
\includegraphics[width=0.97\linewidth]{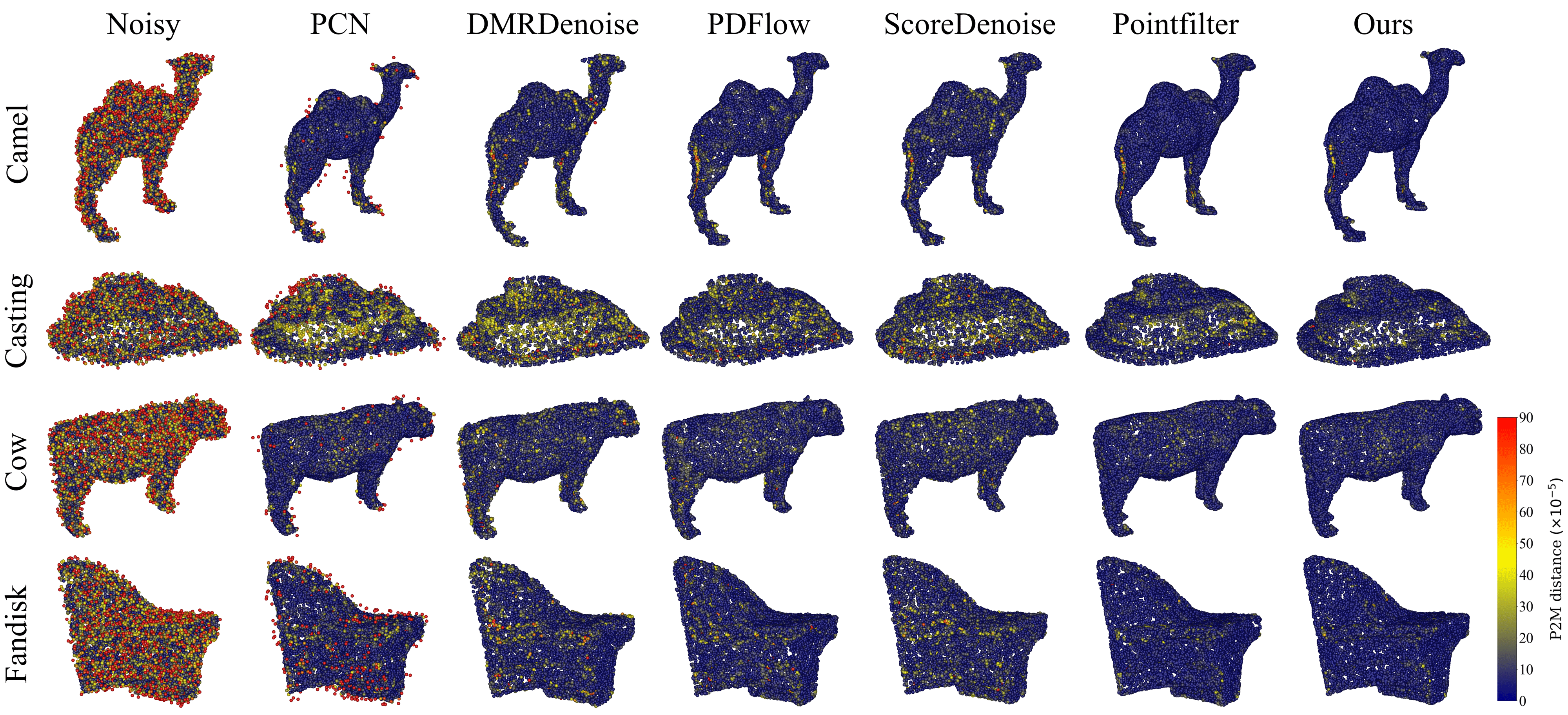}
\caption{Visual results of point-wise P2M distance for 50K resolution shapes with non-isotropic Gaussian noise and a scale parameter of 2\% of the bounding sphere radius. }
\label{fig:all-synthetic-cov-results}
\end{figure*}

In addition to Gaussian noise, we are interested in comparing filtering results for different noise patterns. In this section, we follow the work of ScoreDenoise~\cite{Luo-Score-Based-Denoising} and investigate the filtering performance of methods on the following noise types:

\textbf{Non-isotropic Gaussian noise} where we set the covariance matrix of the Gaussian noise distribution to:
\begin{align}
    \Sigma = s^2\times \begin{bmatrix}
                        1 & -1/2 & -1/4 \\
                        -1/2 & 1 & -1/4 \\
                        -1/4 & -1/4 & 1
                        \end{bmatrix}
\end{align}
The noise scale parameter $s$ is set to $1\%$, $2\%$ and $2.5\%$ of the bounding sphere's radius. Results are presented in Table~\ref{tab:syn-data-cov} and Fig.~\ref{fig:all-synthetic-cov-results}. Much like the case of isotropic Guassian noise, Table~\ref{tab:syn-data} of the main paper, our method outperforms others and generalizes well to this noise pattern.

\begin{figure*}[!tp]
\centering
\includegraphics[width=0.97\linewidth]{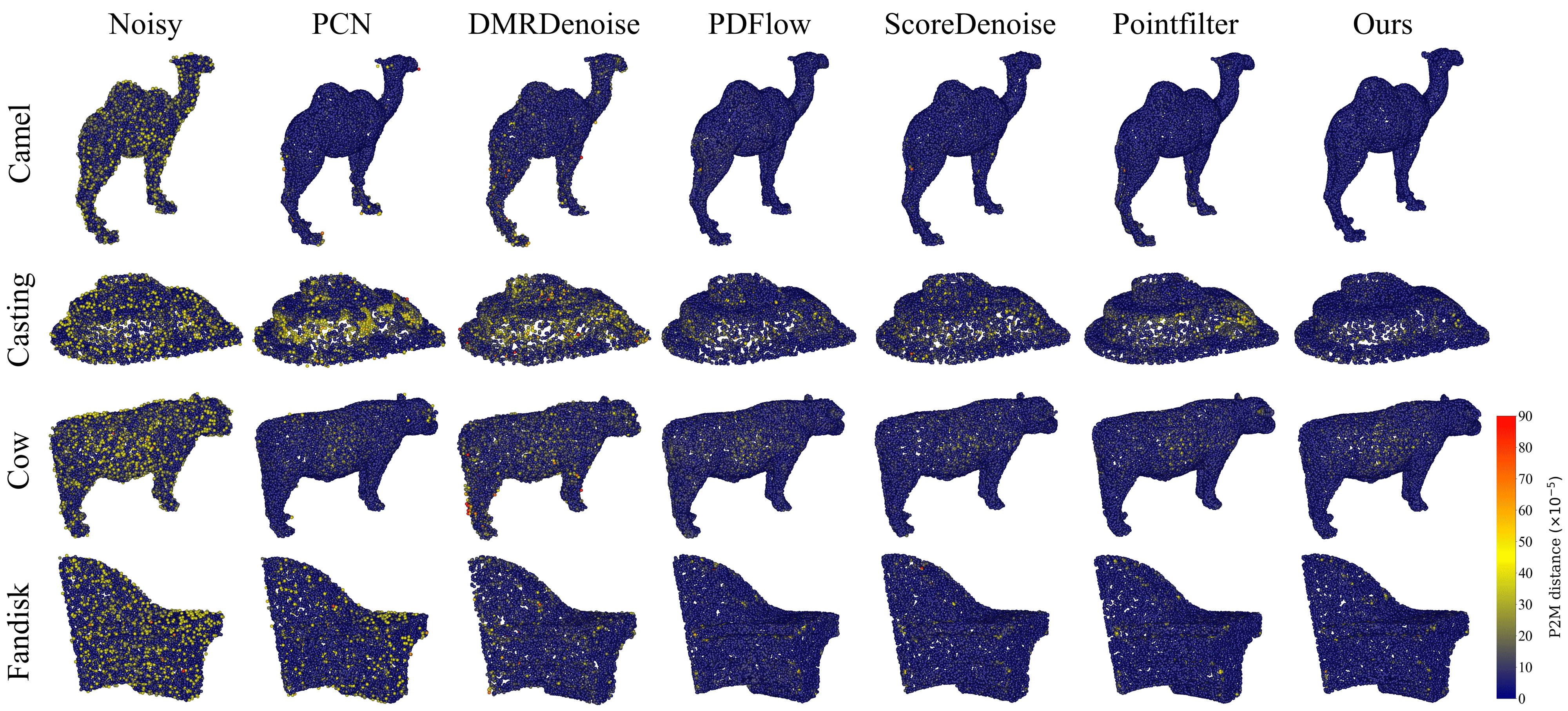}
\caption{Visual results of point-wise P2M distance for 50K resolution shapes with discrete noise and a scale parameter of 2\% of the bounding sphere radius. }
\label{fig:all-synthetic-discrete-results}
\end{figure*}

\textbf{Discrete noise} with the following distribution:
\begin{align}
    p(\pmb{x}; s) = \begin{cases} 0.1, &\pmb{x}=(\pm s, 0, 0)~\text{or}~\pmb{x}=(0, \pm s, 0)\\&~\text{or}~\pmb{x}=(0, 0, \pm s), \\
    0.4, &\pmb{x}=(0, 0, 0) \\ 
    0, &\text{Otherwise}
    \end{cases}
\end{align}
where $s$ is set to $1\%$, $2\%$ and $2.5\%$ of the bounding sphere's radius. Results are presented in Table~\ref{tab:syn-data-discrete} and Fig.~\ref{fig:all-synthetic-discrete-results}. Our method outperforms others across resolutions and noise scales.

\begin{figure*}[!tp]
\centering
\includegraphics[width=0.97\linewidth]{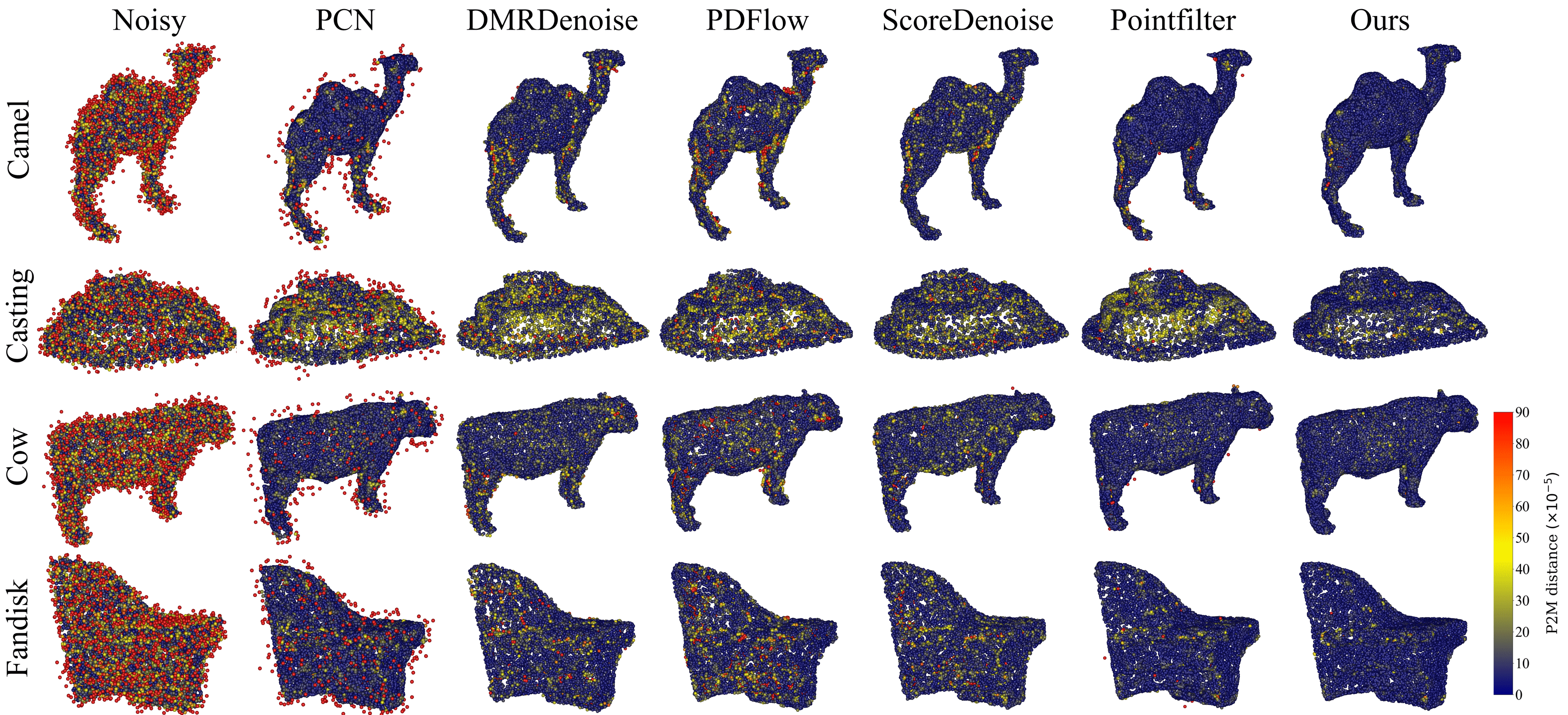}
\caption{Visual results of point-wise P2M distance for 50K resolution shapes with Laplace noise and scale of 2\% of the bounding sphere radius. }
\label{fig:all-synthetic-laplacian-results}
\end{figure*}

\textbf{Laplace noise} with the noise scale set to $1\%$, $2\%$ and $2.5\%$ of the bounding sphere's radius. Results are presented in Table~\ref{tab:syn-data-laplacian} and Fig.~\ref{fig:all-synthetic-laplacian-results}. This noise pattern has a relatively high noise intensity as illustrated by the CD and P2M metric results for the noisy point clouds. Nevertheless, our method consistently outperforms other methods and recovers filtered point clouds with low CD and P2M distances from the clean counterparts.

\begin{figure*}[!tp]
\centering
\includegraphics[width=0.97\linewidth]{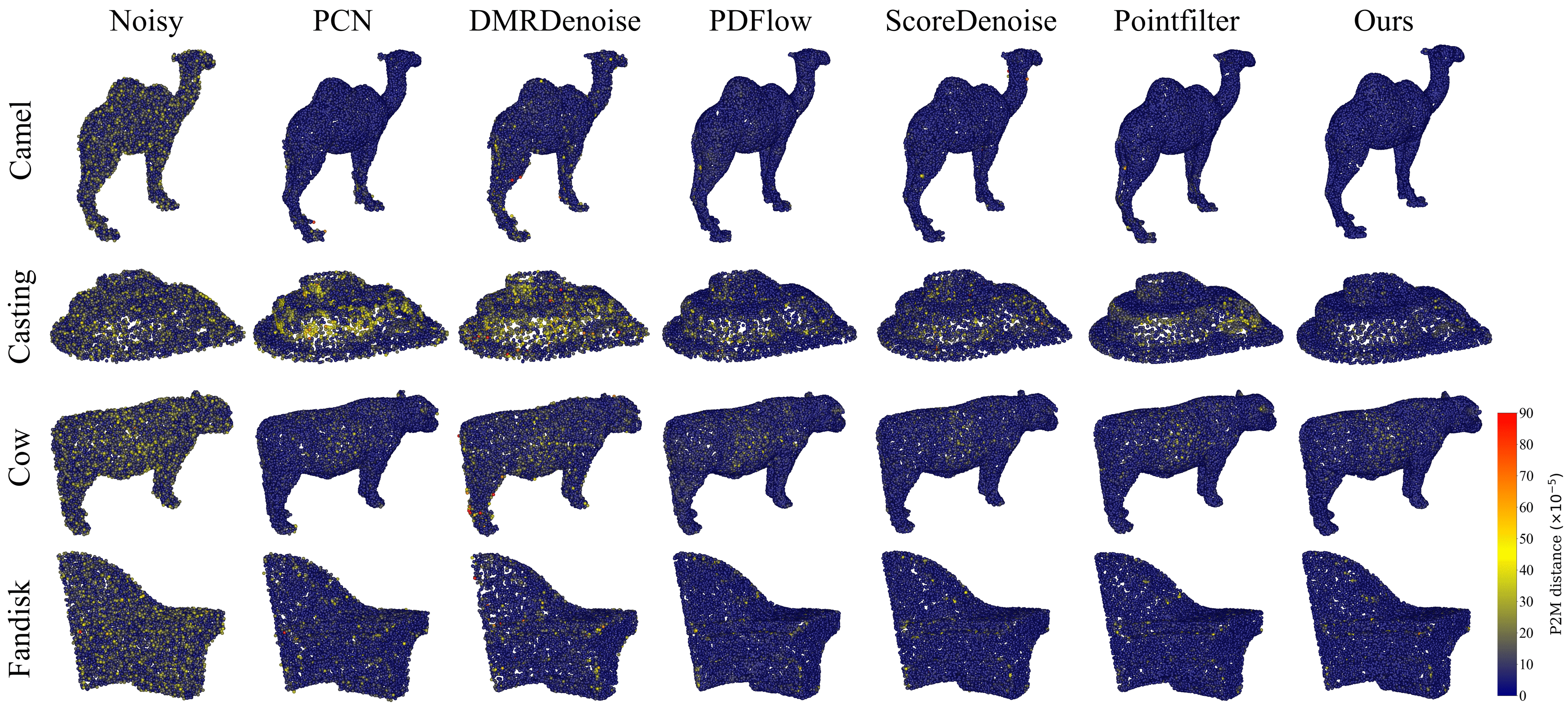}
\caption{Visual results of point-wise P2M distance for 50K resolution shapes with noise uniformly distributed within a 3D sphere of radius $s$. Here, $s$ corresponds to the noise scale and is equal to 2\% of the bounding sphere radius. }
\label{fig:all-synthetic-uniform-results}
\end{figure*}

\textbf{Uniform distribution} of noise within a 3D sphere of radius $s$, given by,
\begin{align}
    p(\pmb{x}; s) = \begin{cases} \frac{3}{4\pi s^3}, &\norm{\pmb{x}}_2\leq s, \\
    0, &\text{Otherwise}
    \end{cases}
\end{align}
where $s$ is set to $1\%$, $2\%$ and $2.5\%$ of the bounding sphere's radius. Results are presented in Table~\ref{tab:syn-data-uniform-ball} and Fig.~\ref{fig:all-synthetic-uniform-results}. This noise pattern corresponds to noise sampled uniformly within a sphere of radius $s$, and as pointed out by~\cite{Luo-Score-Based-Denoising}, is an example of a noise distribution that is not uni-modal and is unlike Gaussian or Laplace noise. Our method again outperforms other state-of-the-art methods.

\midsepremove
\begin{table*}[!tp]
\centering
\begin{tabular}{l|llllll|llllll}
\toprule
\multicolumn{1}{c|}{\multirow{3}{*}{Method}} & \multicolumn{6}{c|}{10K points} & \multicolumn{6}{c}{50K points} \\
\cmidrule{2-13}
 & \multicolumn{2}{c|}{1\% noise} & \multicolumn{2}{c|}{2\% noise} & \multicolumn{2}{c|}{2.5\% noise} & \multicolumn{2}{c|}{1\% noise} & \multicolumn{2}{c|}{2\% noise} & \multicolumn{2}{c}{2.5\% noise} \\
\cmidrule{2-13}
 & CD & P2M & CD & P2M & CD & P2M & CD & P2M & CD & P2M & CD & P2M \\
\midrule
Noisy & 11.6 & 5.87 & 30.08 & 12.73 & 36.99 & 17.35 & 6.71 & 4.44 & 13.76 & 10.31 & 18.36 & 14.56 \\
\midrule
Bilateral & 19.6 & 15.33 & 27.84 & 15.08 & 28.78 & 14.56 & 15.61 & 14.38 & 15.85 & 14.09 & 15.74 & 13.74 \\
Jet & 17.82 & 10.46 & 38.94 & 22.82 & 41.06 & 23.74 & 4.57 & 3.22 & 8.13 & 5.67 & 8.94 & 6.22 \\
WLOP & 59.72 & 29.85 & 66.12 & 35.0 & 68.2 & 36.57 & 8.28 & 5.31 & 10.11 & 6.77 & 11.32 & 7.73 \\
\midrule
PCN & 11.64 & 5.89 & 30.06 & 12.71 & 36.95 & 17.31 & 6.21 & 4.14 & 8.86 & 5.95 & 10.26 & 7.05 \\
GPDNet & 7.88 & 4.33 & 19.19 & 6.34 & 22.32 & 7.65 & 4.6 & 3.14 & 7.75 & 5.29 & 10.38 & 7.45 \\
DMRDenoise & 48.4 & 24.25 & 48.42 & 23.74 & 48.43 & 23.86 & 11.1 & 6.9 & 12.5 & 7.95 & 12.68 & 8.13 \\
PDFlow & 8.71 & 4.64 & 18.86 & 6.33 & 23.37 & 8.29 & 4.33 & 3.04 & 6.72 & 4.53 & 7.38 & 5.0 \\
ScoreDenoise & 12.44 & 5.32 & 21.82 & 7.07 & 28.81 & 11.72 & 4.48 & 2.91 & 6.16 & 3.87 & 7.32 & 4.62 \\
Pointfilter & 11.14 & 5.72 & 20.56 & 6.92 & 22.49 & 7.57 & 5.43 & 3.92 & 6.12 & 4.16 & 6.43 & 4.34 \\
\midrule
\textbf{Ours} & \textbf{6.41} & \textbf{3.67} & \textbf{16.49} & \textbf{4.66} & \textbf{18.72} & \textbf{5.21} & \textbf{3.47} & \textbf{2.53} & \textbf{4.26} & \textbf{2.8} & \textbf{4.62} & \textbf{2.99}
\\
\bottomrule
\end{tabular}
\caption{Filtering results on the PUNet dataset with discrete noise. CD and P2M distances are multiplied by $10^5$.}
\label{tab:syn-data-discrete} 
\end{table*}
\midsepdefault

\midsepremove
\begin{table*}[!tp]
\centering
\setlength\tabcolsep{5pt} % default value: 6pt
\begin{tabular}{l|llllll|llllll}
\toprule
\multicolumn{1}{c|}{\multirow{3}{*}{Method}} & \multicolumn{6}{c|}{10K points} & \multicolumn{6}{c}{50K points} \\
\cmidrule{2-13}
 & \multicolumn{2}{c|}{1\% noise} & \multicolumn{2}{c|}{2\% noise} & \multicolumn{2}{c|}{2.5\% noise} & \multicolumn{2}{c|}{1\% noise} & \multicolumn{2}{c|}{2\% noise} & \multicolumn{2}{c}{2.5\% noise} \\
\cmidrule{2-13}
 & CD & P2M & CD & P2M & CD & P2M & CD & P2M & CD & P2M & CD & P2M \\
\midrule
Noisy & 49.83 & 25.95 & 117.73 & 83.79 & 162.23 & 124.83 & 28.69 & 22.23 & 86.64 & 76.74 & 127.21 & 115.91 \\
\midrule
Bilateral & 36.87 & 17.95 & 80.15 & 51.36 & 112.47 & 78.99 & 18.2 & 14.66 & 41.25 & 35.58 & 71.13 & 63.96 \\
Jet & 35.87 & 15.64 & 63.13 & 38.0 & 70.8 & 44.24 & 9.4 & 5.43 & 18.63 & 13.04 & 28.33 & 21.79 \\
WLOP & 60.94 & 31.19 & 103.49 & 69.38 & 126.41 & 88.96 & 11.73 & 7.98 & 32.5 & 25.96 & 52.69 & 44.16 \\
\midrule
PCN & 49.75 & 25.87 & 117.6 & 83.64 & 162.1 & 124.66 & 13.71 & 8.89 & 47.62 & 40.94 & 86.11 & 78.13 \\
GPDNet & 28.61 & 10.13 & 63.79 & 36.67 & 96.58 & 65.43 & 15.33 & 10.53 & 58.14 & 49.37 & 93.35 & 82.92 \\
DMRDenoise & 48.45 & 23.48 & 55.12 & 28.84 & 59.61 & 32.72 & 12.6 & 8.12 & 17.62 & 12.62 & 23.45 & 17.98 \\
PDFlow & 25.38 & 8.76 & 43.19 & 22.05 & 51.87 & 30.12 & 8.21 & 5.52 & 22.76 & 17.89 & 45.08 & 38.08 \\
ScoreDenoise & 29.03 & 9.64 & 46.02 & 21.36 & 51.84 & 27.3 & 8.25 & 4.89 & 16.77 & 11.67 & 18.51 & 12.79 \\
Pointfilter & 27.73 & 8.61 & 42.5 & 16.73 & 54.66 & 25.55 & 8.27 & 4.85 & 12.44 & 7.59 & 17.54 & 11.69 \\
\midrule
\textbf{Ours} & \textbf{23.93} & \textbf{6.02} & \textbf{33.96} & \textbf{11.0} & \textbf{42.93} & \textbf{17.55} & \textbf{6.53} & \textbf{3.36} & \textbf{9.99} & \textbf{5.8} & \textbf{16.6} & \textbf{10.96} \\
\bottomrule
\end{tabular}
\caption{Filtering results on the PUNet dataset with Laplace noise. CD and P2M distances are multiplied by $10^5$.}
\label{tab:syn-data-laplacian} 
\end{table*}
\midsepdefault

\midsepremove
\begin{table*}[!tp]
\centering
\begin{tabular}{l|llllll|llllll}
\toprule
\multicolumn{1}{c|}{\multirow{3}{*}{Method}} & \multicolumn{6}{c|}{10K points} & \multicolumn{6}{c}{50K points} \\
\cmidrule{2-13}
 & \multicolumn{2}{c|}{1\% noise} & \multicolumn{2}{c|}{2\% noise} & \multicolumn{2}{c|}{2.5\% noise} & \multicolumn{2}{c|}{1\% noise} & \multicolumn{2}{c|}{2\% noise} & \multicolumn{2}{c}{2.5\% noise} \\
\cmidrule{2-13}
 & CD & P2M & CD & P2M & CD & P2M & CD & P2M & CD & P2M & CD & P2M \\
\midrule
Noisy & 11.89 & 6.21 & 35.01 & 14.17 & 44.54 & 19.62 & 7.96 & 4.7 & 16.93 & 10.99 & 22.45 & 15.55 \\
\midrule
Bilateral & 19.92 & 15.42 & 31.4 & 14.78 & 33.86 & 14.46 & 16.49 & 14.41 & 17.25 & 14.14 & 17.01 & 13.6 \\
Jet & 18.15 & 10.55 & 42.27 & 22.72 & 46.06 & 24.38 & 5.6 & 3.28 & 9.56 & 5.8 & 10.42 & 6.4 \\
WLOP & 59.88 & 29.79 & 71.06 & 37.63 & 67.11 & 35.71 & 8.12 & 5.17 & 10.63 & 7.2 & 12.05 & 8.22 \\
\midrule
PCN & 11.93 & 6.22 & 34.96 & 14.14 & 44.48 & 19.56 & 7.24 & 4.25 & 10.62 & 6.09 & 11.81 & 6.92 \\
GPDNet & 7.84 & 4.31 & 22.31 & 6.59 & 27.25 & 8.22 & 5.45 & 3.24 & 9.97 & 5.87 & 13.66 & 8.6 \\
DMRDenoise & 49.14 & 24.01 & 47.69 & 22.61 & 48.06 & 22.67 & 11.18 & 6.9 & 12.23 & 7.71 & 12.19 & 7.72 \\
PDFlow & 8.74 & 4.64 & 20.26 & 6.24 & 24.65 & 8.38 & 4.56 & 3.03 & 6.82 & 4.54 & 7.55 & 5.09 \\
ScoreDenoise & 12.74 & 5.35 & 24.67 & 7.01 & 31.4 & 11.68 & 5.05 & 2.88 & 6.91 & 3.78 & 8.02 & 4.74 \\
Pointfilter & 11.39 & 5.7 & 24.47 & 6.92 & 27.76 & 7.68 & 6.31 & 3.96 & 7.42 & 4.2 & 7.63 & 4.23 \\
\midrule
\textbf{Ours} & \textbf{6.45} & \textbf{3.68} & \textbf{20.09} & \textbf{4.74} & \textbf{24.16} & \textbf{5.45} & \textbf{4.43} & \textbf{2.54} & \textbf{5.99} & \textbf{2.98} & \textbf{6.42} & \textbf{3.24} \\
\bottomrule
\end{tabular}
\caption{Filtering results on the PUNet dataset with noise uniformly distributed within a 3D sphere of radius $s$. Here, $s$ corresponds to the noise scale. CD and P2M distances are multiplied by $10^5$.}
\label{tab:syn-data-uniform-ball} 
\end{table*}
\midsepdefault

\midsepremove
\begin{table}[!tp]
\centering
\setlength\tabcolsep{3pt} % default value: 6pt
\begin{tabular}{c|ll|ll|ll}
\toprule
\multirow{2}{*}{\shortstack{Ablation:\\10K points}} & \multicolumn{2}{c|}{2.75\% noise} & \multicolumn{2}{c|}{3\% noise} & \multicolumn{2}{c}{3.25\% noise} \\
\cmidrule{2-7}
& CD & P2M & CD & P2M & CD & P2M \\ 
\midrule
$\mathcal{L}_a$ \& 1 it. & 40.76 & 15.46 & 46.33 & 19.62 & 53.55 & 25.36 \\
$\mathcal{L}_a$ \& 2 it. & 37.53 & 13.2 & 43.91 & 17.82 & 52.8 & 24.72 \\
\textbf{$\mathcal{L}_a$ \& 4 it.} & \textbf{36.31} & \textbf{12.3} & \textbf{41.87} & \textbf{16.44} & \textbf{51.2} & \textbf{23.6} \\
$\mathcal{L}_a$ \& 8 it. & 37.44 & 13.15 & 45.39 & 19.1 & 56.65 & 27.97 \\
$\mathcal{L}_a$ \& 12 it. & 37.95 & 13.59 & 45.14 & 18.86 & 55.17 & 26.79 \\
$\mathcal{L}_a$ \& DPFN & 39.49 & 14.48 & 47.06 & 20.11 & 57.35 & 28.25  \\
\midrule
$\mathcal{L}_b$ \& 4 it. & 37.62 & 13.2 & 44.22 & 18.13 & 54.36 & 25.93 \\
\bottomrule
\end{tabular}
\caption{\dt{Ablation results for different iteration numbers and loss functions at higher noise. CD and P2M distances are multiplied by $10^5$.}}
\label{tab:ablation-high-noise} 
\end{table}
\midsepdefault

\dt{
\section{Further Ablations on Iteration Number}
\label{sec:ablation-it-number}
In Table~\ref{tab:ablation-high-noise} we consider the impact of iteration number under higher noise settings. The maximum training noise for all models is $2\%$, much lower than the unseen, high noise settings of Table~\ref{tab:ablation-high-noise}. We observe that the optimum number of filtering iterations for our network is 4. Variants containing higher numbers of ItMs appear to over-specialize on the training noise scales and perform sub-optimally at higher, unseen noise scales. Moreover, the DPFN network, with 1 ItM comprising 6 Dynamic EdgeConv layers as opposed to 4 ItMs with 4 Dynamic EdgeConv layers each, performs poorly with increasing noise. This result demonstrates the importance of multiple ItMs and \textit{true} iterative filtering, as opposed to a deep, 1 iteration, network with the same number of parameters. }

\begin{figure}[!tp]
\centering
\includegraphics[width=0.47\textwidth]{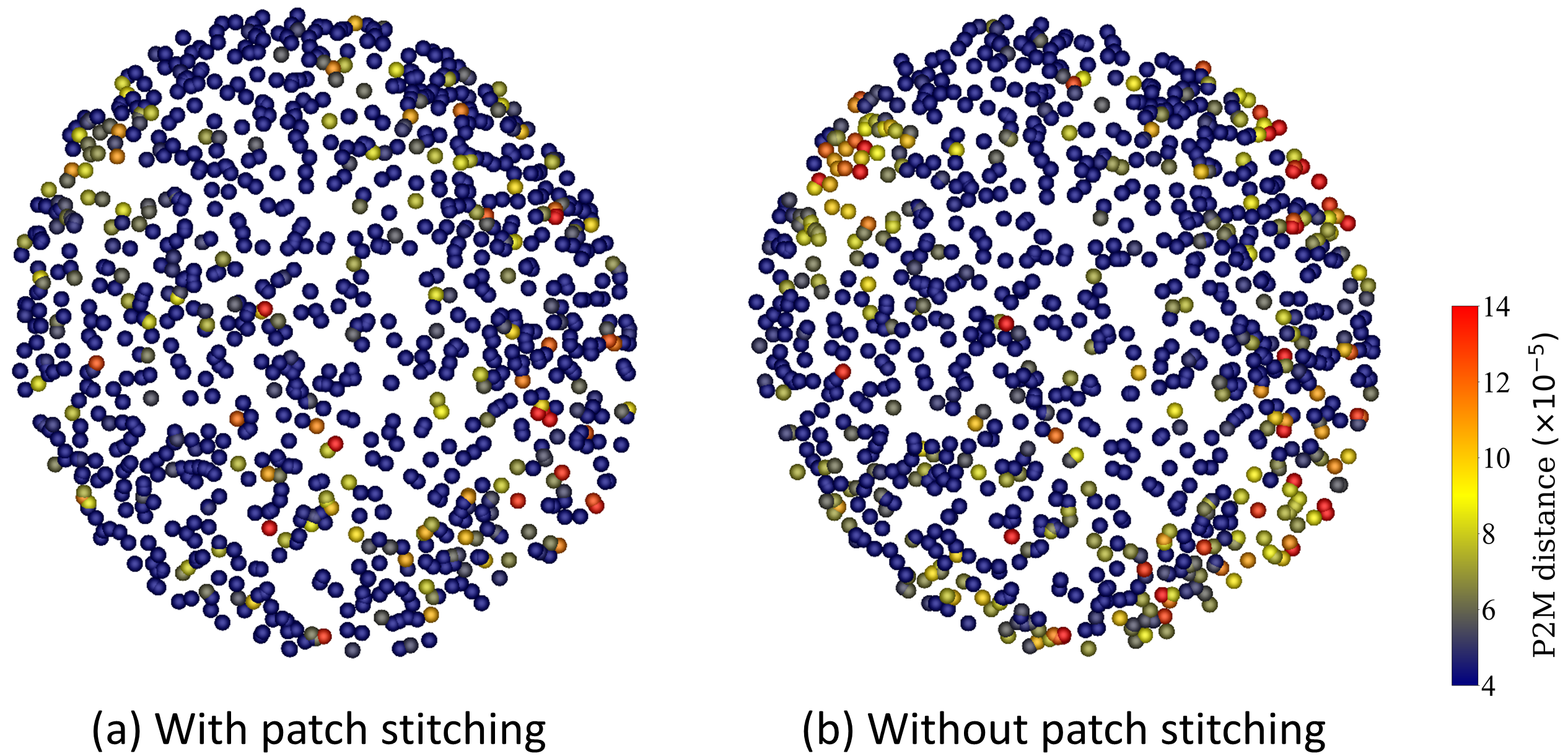}
\caption{Patch (a) illustrates a noisy patch filtered with patch stitching and patch (b) corresponds to the same noisy patch filtered without patch stitching. The point-wise P2M distance is given for each point. As shown by patch (b), we observe that boundary points of patches tend to have higher P2M distances, i.e., filtered results away from the center are less accurate, when filtering without patch stitching. However, results closer to the center have higher accuracy. Therefore, patch stitching attempts to recover the best filtered points from overlapping patches and reduce the filtering error at boundary points. This can be seen in patch (a), where stitching is incorporated. We observe more accurate filtering results near the boundary.}
\label{fig:patch-stitching}
\end{figure}

\section{Ablation Study on Patch Stitching}
\label{sec:ablation-patch-stitching}
During inference, given a noisy point cloud, we construct patches of 1000 nearest neighbors from a set of reference points $\{\pmb{x}_r\}^R_{r=1}$ determined using farthest point sampling. Typically, for a noisy point cloud of 50K points, we construct 300 such patches and filter each patch with all patch points being simultaneously filtered. As these 300 patches have overlapping regions, some points may be repeated amongst multiple patches which leads to some noisy points having multiple filtering results associated to them. Therefore, it is necessary to select the best filtered points that correspond to the best filtering result for each point amongst all patches. To achieve this, inspired by Zhou et al.~\cite{Zhou-Patch-Stitching}, we design a generalized patch stitching mechanism to utilize Gaussian weights when selecting best filtered points. These weights are designed such that noisy points closer to a patch's central reference point, $\pmb{x}_r$, are weighted higher than points further away from it. This is motivated by the observation that points at the patch boundary are filtered less favorably compared to those close to the central point. During inference, we recover the best filtered point, corresponding to a given initial noisy point, by selecting the filtered result within the patch where the noisy point was weighted highest. 

\midsepremove
\begin{table}[!tp]
\centering
\setlength\tabcolsep{5pt} % default value: 6pt
\begin{tabular}{c|ll|ll|ll}
\toprule
\multirow{3}{*}{Ablation} & \multicolumn{6}{c}{10K points} \\
\cmidrule{2-7}
& \multicolumn{2}{c|}{1\% noise} & \multicolumn{2}{c|}{2\% noise} & \multicolumn{2}{c}{2.5\% noise} \\
\cmidrule{2-7}
& CD & P2M & CD & P2M & CD & P2M \\ 
\midrule
without PS & 21.19 & 5.45 & 32.38 & 10.2 & 38.67 & 14.98 \\
\midrule
\textbf{with PS} & \textbf{20.56} & \textbf{5.01} & \textbf{30.43} & \textbf{8.45} & \textbf{33.52} & \textbf{10.45} \\
\bottomrule
\end{tabular}
\caption{Ablation results with and without patch stitching (PS). CD and P2M distances are multiplied by $10^5$.}
\label{tab:ablation-ps} 
\end{table}
\midsepdefault 

\midsepremove
\begin{table*}[!tp]
\setlength{\tabcolsep}{5pt}
\centering
\begin{tabular}{l|llllll|llllll}
\toprule
\multicolumn{1}{c|}{\multirow{3}{*}{Method}} & \multicolumn{6}{c|}{10K points} & \multicolumn{6}{c}{50K points} \\
\cmidrule{2-13}
 & \multicolumn{2}{c|}{1\% noise} & \multicolumn{2}{c|}{2\% noise} & \multicolumn{2}{c|}{2.5\% noise} & \multicolumn{2}{c|}{1\% noise} & \multicolumn{2}{c|}{2\% noise} & \multicolumn{2}{c}{2.5\% noise} \\
\cmidrule{2-13}
 & CD & P2M & CD & P2M & CD & P2M & CD & P2M & CD & P2M & CD & P2M \\
\midrule
Noisy & 36.9 & 16.03 & 79.39 & 47.72 & 105.02 & 70.03 & 18.69 & 12.82 & 50.48 & 41.36 & 72.49 & 62.03 \\
Ours (PN++) & 22.9 & 5.93 & 33.07 & 10.28 & 36.92 & 13.06 & 6.18 & 3.09 & 9.23 & 5.22 & 12.1 & 7.47 \\
Ours (PN++ w/ GRU) & 22.81 & 5.81 & 33.15 & 10.31 & 37.08 & 13.17 & 6.15 & 3.07 & 9.19 & 5.2 & 12.04 & 7.44 \\
\midrule
\textbf{Ours} & \textbf{20.56} & \textbf{5.01} & \textbf{30.43} & \textbf{8.45} & \textbf{33.52} & \textbf{10.45} & \textbf{6.05} & \textbf{3.02} & \textbf{8.03} & \textbf{4.36} & \textbf{10.15} & \textbf{5.88} \\
\bottomrule
\end{tabular}
\caption{\dt{Filtering results on the PUNet dataset for IterativePFNs with Pointnet++ based encoders as compared to the original graph convolution based implementation. CD and P2M distances are multiplied by $10^5$.}}
\label{tab:syn-data-diff-backbones} 
\end{table*}
\midsepdefault

The impact of incorporating patch stitching can be seen in Fig.~\ref{fig:patch-stitching}. Here, patch (b) is filtered without patch stitching. We see that many boundary points have high point-wise P2M values, indicating these filtered results lie further away from the clean surface. By contrast, patch (a), filtered using patch stitching, has fewer boundary points with high P2M values. Table~\ref{tab:ablation-ps} shows that patch stitching leads to better overall filtering, especially at high noise scales.

\dt{
\section{Filtering Results for PointNet++ based Encoders}
\label{sec:ablation-different-backbones}
We are motivated that filtering is the reverse Markov process that iteratively removes noise. To obtain the filtered point $\pmb{x}^{\tau}_i$, we only need the filtered point from the previous iteration, i.e., $\pmb{x}^{\tau-1}_i$. Hence, a simple Encoder-Decoder  module (ItM) suffices to obtain filtered displacements. To test our hypothesis, and demonstrate the generalization of our method to other 3D point set convolution architectures, we create variant IterativePFNs with 4 ItMs composed of Pointnet++ based encoders: 
\begin{itemize}
    \item \textbf{PN++}: a vanilla Pointnet++ based encoder where the graph convolution of point cloud patches, of our original implementation, is replaced by direct point set convolution of these patches using a Pointnet++ achitecture. 
    \item \textbf{PN++ w/ GRU}: similar to~\cite{Wen-PMPNet}, we create a Pointnet++ based variant which incorporates GRU layers within the encoder to maintain an additional \textit{memory} of feature information from the previous iteration.  
\end{itemize} 
Our original network, which uses a DGCNN-based encoder, performs better as shown in Table~\ref{tab:syn-data-diff-backbones}. Using an additional GRU layer within the encoder provides no meaningful performance gain as the results of the vanilla Pointnet++ implementation performs comparably to the PN++ w/ GRU network. This reinforces our hypothesis that filtering a point, $\pmb{x}^{\tau}_i$, at the subsequent iteration, is a reverse Markov process which requires only the filtered point, $\pmb{x}^{\tau-1}_i$, from the preceding iteration.
}

\section{Runtimes for Learning based Methods}
\label{sec:runtimes}
Finally, we compare runtimes across different learning based methods. We do not compare conventional methods as they usually require parameter tuning and, therefore, user interaction, to obtain best results. Furthermore, the bilateral filtering mechanism, and the upsampling step for WLOP, both require point normals to be computed separately. This process can be quite time-consuming. Our method is quite competitive, ranking third in runtime while also achieving best results on synthetic and scanned data. By contrast, DMRDenoise~\cite{Luo-DMRDenoise}, which has the shortest runtime, generally performs poorly on the filtering task. ScoreDenoise~\cite{Luo-Score-Based-Denoising}, while performing relatively faster, still produces sub-optimal filtering results. The ScoreDenoise runtime mentioned in Table~\ref{tab:runtimes} corresponds to filtered results after the default number of 30 gradient ascent steps, i.e., 30 test time iterations. However, as filtered results retain significant amounts of noise, additional iterations will be required to further decrease this noise which inevitably contributes to higher runtimes. GPDNet, Pointfilter and PCN perform worst in terms of efficiency. In the case of Pointfilter and PCN, the one patch per central point filtering strategy contributes to these high runtimes since each input patch is used to filter a single central point. By contrast, our method filters all patch points simultaneously while performing iterative filtering internally and, coupled with the generalized patch stitching mechanism we introduce, yields the best filtering results with short runtimes.

\midsepremove
\begin{table}[!tp]
\centering
\begin{tabular}{l|ll|ll}
\toprule
Method & Time (s) \\
\midrule
PCN & 167.75 \\
GPDNet & 79.87 \\
DMRDenoise & \textbf{11.08} \\
PDFlow & 32.74\\
ScoreDenoise &  18.7\\
Pointfilter & 84.32 \\
\midrule
Ours & 22.91 \\
\bottomrule
\end{tabular}
\caption{Runtimes of different learning based methods for filtering a noisy point cloud of 50K points with 2\% Gaussian noise.}
\label{tab:runtimes} 
\end{table}
\midsepdefault

\end{document}